\documentclass[10pt,journal,compsoc]{IEEEtran}

\usepackage{makecell}
\usepackage{hyperref}
\usepackage{dblfloatfix} 
\usepackage{float}
\usepackage{hyperref}
\usepackage[T1]{fontenc}
\usepackage[utf8]{inputenc}
\usepackage{graphicx} 
\usepackage{booktabs} 
\usepackage{multirow}
\usepackage{array}
\usepackage{amssymb}
\usepackage{amsmath}
\usepackage{pdflscape}
\usepackage{adjustbox}
\usepackage{rotating}
\usepackage{afterpage}
\usepackage{multirow}
\usepackage{placeins}
\usepackage{xcolor}

\ifCLASSOPTIONcompsoc
  \usepackage[nocompress]{cite}
\else
  \usepackage{cite}
\fi
\ifCLASSINFOpdf
\else
\fi
\begin{document}
%
\title{A Review of Human Emotion Synthesis Based on Generative Technology}
%
%
%
%

\author{Fei Ma*,
        Yukan Li*,
        Yifan Xie*,
        Ying He,
        Yi Zhang,
        Hongwei Ren,
        Zhou Liu,
        Wei Yao, 
        Fuji Ren,~\IEEEmembership{Senior Member,~IEEE},
        Fei Richard Yu,~\IEEEmembership{Fellow,~IEEE},
        Shiguang Ni
\IEEEcompsocitemizethanks{
\IEEEcompsocthanksitem (Corresponding Author: Shiguang Ni. *: Equal Contribution.)
\IEEEcompsocthanksitem Fei Ma, Yifan Xie, Yi Zhang, and Zhou Liu are with Guangdong Laboratory of Artificial Intelligence and Digital Economy (SZ), Shenzhen, China.
E-mail: mafei@gml.ac.cn, xieyifan@stu.xjtu.edu.cn, zhangyi@gml.ac.cn, liuzhou@gml.ac.cn.
\IEEEcompsocthanksitem Yukan Li, Wei Yao, and Shiguang Ni are with Tsinghua Shenzhen International Graduate School, Tsinghua University, Shenzhen, China.
E-mail: liyukan23@mails.tsinghua.edu.cn, viviayao@sz.tsinghua.edu.cn, ni.shiguang@sz.tsinghua.edu.cn. 
\IEEEcompsocthanksitem Ying He is with College of Computer Science and Software Engineering, Shenzhen University, Shenzhen, China.
E-mail: heying@szu.edu.cn.
\IEEEcompsocthanksitem Hongwei Ren is with MICS Thrust, The Hong Kong University of Science and Technology (GZ), Guangzhou, China.
E-mail: hren066@connect.hkust-gz.edu.cn.
\IEEEcompsocthanksitem Fuji Ren is with School of Computer Science and Engineering, University of Electronic Science and Technology of China, Chengdu, China, and also with Shenzhen Institute for Advanced Study, University of Electronic Science and Technology of China, Shenzhen, China.
E-mail: renfuji@uestc.edu.cn.
\IEEEcompsocthanksitem Fei Richard Yu is with College of Computer Science and Software Engineering, Shenzhen University, Shenzhen, China, and also with School of Information Technology, Carleton University, Canada.
E-mail: richard.yu@ieee.org.
}
\thanks{Manuscript received April 19, 2005; revised August 26, 2015.}}

%
%

\markboth{Journal of \LaTeX\ Class Files,~Vol.~14, No.~8, August~2015}%
{Shell \MakeLowercase{\textit{et al.}}: Bare Advanced Demo of IEEEtran.cls for IEEE Computer Society Journals}
%



\IEEEtitleabstractindextext{%
\begin{abstract}
Human emotion synthesis is a crucial aspect of affective computing. It involves using computational methods to mimic and convey human emotions through various modalities, with the goal of enabling more natural and effective human-computer interactions. 
Recent advancements in generative models, such as Autoencoders, Generative Adversarial Networks, Diffusion Models, Large Language Models, and Sequence-to-Sequence Models, have significantly contributed to the development of this field. 
However, there is a notable lack of comprehensive reviews in this field.
To address this problem, this paper aims to address this gap by providing a thorough and systematic overview of recent advancements in human emotion synthesis based on generative models. 
Specifically, this review will first present the review methodology, the emotion models involved, the mathematical principles of generative models, and the datasets used.
Then, the review covers the application of different generative models to emotion synthesis based on a variety of modalities, including facial images, speech, and text.
It also 
examines mainstream evaluation metrics. 
Additionally, the review presents some major findings 
and suggests future research directions, providing a comprehensive understanding of the role of generative technology in the nuanced domain of emotion synthesis.
\end{abstract}

}

\maketitle

\IEEEdisplaynontitleabstractindextext

%
\IEEEpeerreviewmaketitle

\ifCLASSOPTIONcompsoc
\IEEEraisesectionheading{\section{Introduction}\label{sec:introduction}}
\else
\section{Introduction}
\label{sec:introduction}
\fi

%
%
%
%

\IEEEPARstart{A}{ffective} computing is an interdisciplinary research field that aims to endow computers with the ability to recognize, understand, express, and respond to human emotions \cite{picard2000affective,10131997}. 
It integrates theories and methods from multiple disciplines such as computer science, psychology, and cognitive science, attempting to reveal the essence of human emotions and apply it to human-computer interaction and intelligent systems. 
The core goal of affective computing is to enable computers to perceive, understand, and express emotions like humans, thereby achieving more natural and friendly human-computer interaction \cite{yadegaridehkordi2019affective,9852364,9330790,10066197}.

Emotion synthesis \cite{schroder2001emotional} is an important branch of affective computing, which aims to enable computers to generate emotional expressions similar to human emotions.
This ability can be realized through various common modalities, such as facial images, speech, and text. 
To achieve emotion synthesis, researchers have proposed a series of traditional methods by analyzing the characteristics of human emotional expressions and establishing mathematical models. These models are then used to generate speech and facial expressions with specific emotions using computers \cite{tokuda2013speech,hajarolasvadi2020generative}.

Artificial intelligence has made remarkable advances in synthesizing human emotions, marking a significant breakthrough in the field.
In particular, generative technology has greatly improved the effect and application scope of emotion synthesis \cite{poria2017review,cao2023comprehensive,ma2024generative}. 
Compared with traditional methods, these new models can automatically learn the characteristics of emotional expressions from massive data without relying on manually designed rules and models \cite{ruthotto2021introduction,gillioz2020overview,feuerriegel2024generative,xie2024ccis}. 
With their powerful generation capabilities, generative models can generate emotional samples that are highly similar to real data and more flexible, greatly expanding the research boundaries in the field of emotion synthesis.
For example, some researchers use Autoencoders (AEs) \cite{hinton2006reducing} to generate speech with emotions. By modifying this structure, they can extract speaker embeddings, isolate timbre information, and control the flow of emotional attributes \cite{zhang2023iemotts}.
Other researchers use Generative Adversarial Networks (GANs) \cite{goodfellow2014generative} to generate facial images with specific emotions. 
By controlling the input of the generative model, they can generate faces expressing different emotions such as happiness, sadness, and anger \cite{pumarola2018ganimation}. 
In recent years, Diffusion Models (DMs) \cite{ho2020denoising} and Large Language Models (LLMs) \cite{vaswani2017attention} have also been widely used in emotion synthesis tasks. 
Some researchers use DMs to enhance image and audio processing by employing a reconstruction Module, which leverages noising and denoising in latent spaces \cite{zhang2024emotalker}. 
Other researchers use LLMs to generate empathic conversational texts by training language models on empathic conversations and injecting emotional information into response generation \cite{casas2021enhancing}.
However, to the best of our knowledge, there is a conspicuous absence of a systematic review that specifically focuses on generative technology for human emotion synthesis within this burgeoning field.

\begin{figure}[t]
	\centerline{\includegraphics[width=1.0\linewidth]{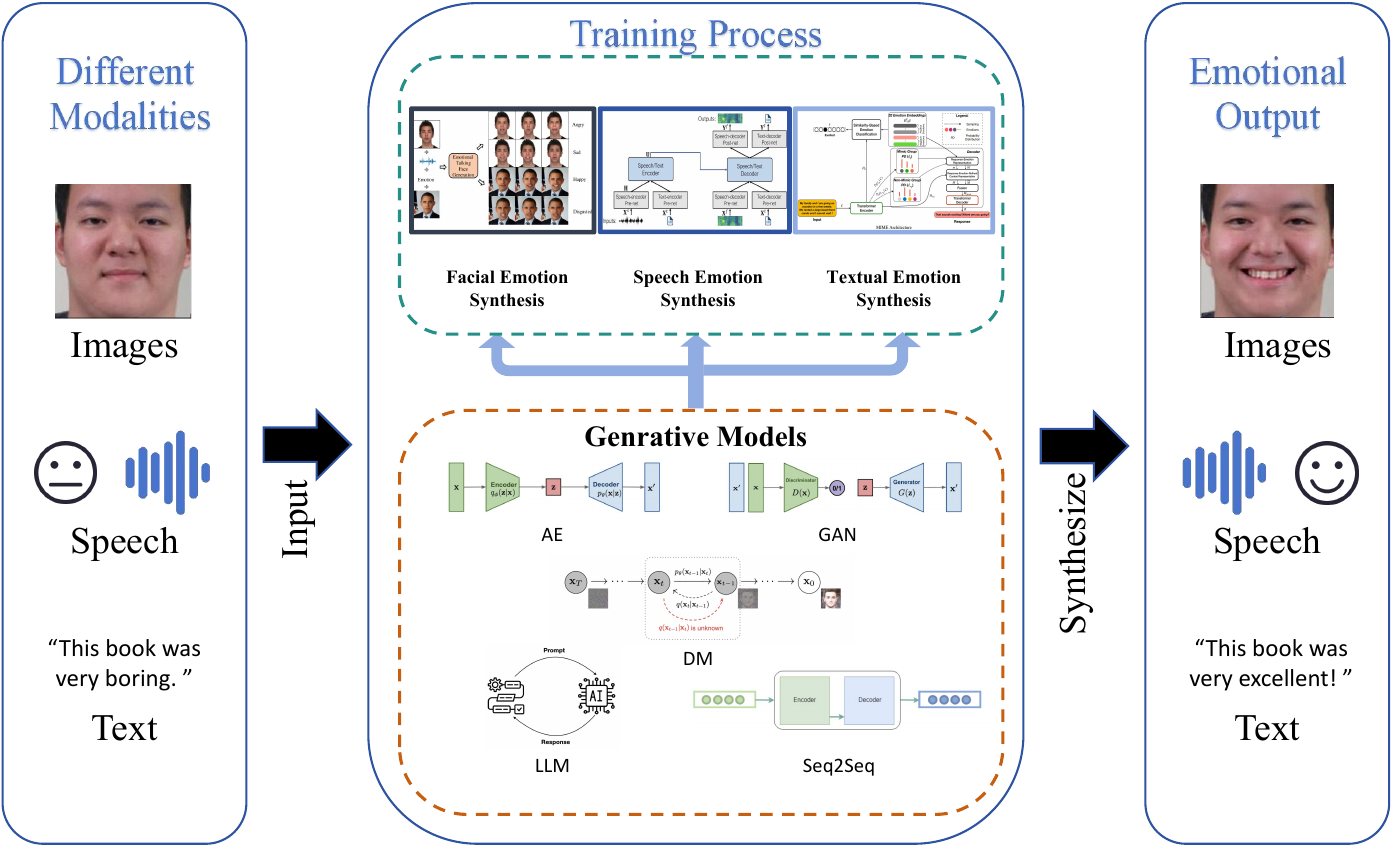}}
    \caption{Schematic Diagram of Generation Technology for Human Emotion Synthesis.}
\scriptsize 
(1) Schematic diagram of facial emotion synthesis is sourced from \url{https://www.semanticscholar.org/reader/2ebb1cb387a1761b20b97c7c3038a7ab119b54d7}, 

(2) Schematic diagram of speech emotion synthesis is sourced from \url{https://huggingface.co/learn/audio-course/chapter6/pre-trained_models}, 

(3) Schematic diagram of speech emotion synthesis is sourced from \url{https://blog.reachsumit.com/posts/2020/12/generating-empathetic-responses/}.

(4) Schematics of GAN and VAE are sourced from \url{https://www.researchgate.net/figure/Comparison-of-two-categories-of-generative-models_fig2_378846405}, 

(5) Seq2Seq schematic is sourced from \url{https://www.geeksforgeeks.org/self-attention-in-nlp/}, 

(6) Diffusion model schematic is sourced from \url{https://bobondemon.github.io/2024/07/18/%E7%B4%80%E9%8C%84-Evidence-Lower-BOund-ELBO-%E7%9A%84%E4%B8%89%E7%A8%AE%E7%94%A8%E6%B3%95/}, 

(7) LLM schematic is sourced from \url{https://events.afcea.org/Augusta24/Custom/Handout/Speaker0_Session11206_1.pdf}.

    \label{fig:figure1}
\end{figure}

This study examines how generative AI models synthesize human emotions, addressing current gaps in research through a systematic analysis.
The overall schematic diagram is illustrated in Fig.~\ref{fig:figure1}. 
Specifically, this paper will firstly introduce several generative models and their underlying mathematical principles to help readers better understand the technical background of this field, including AEs, GANs, DMs, LLMs, and Sequence-to-Sequence (Seq2Seq) models \cite{sutskever2014sequence}. 
AEs learn compressed data representations through encoder-decoder reconstruction, enabling dimensionality reduction and feature extraction. 
GANs use adversarial training between a generator and discriminator to produce realistic data samples. 
DMs generate data by learning to reverse a noise diffusion process, often avoiding vanishing gradients. 
LLMs leverage transformer architectures and massive text datasets for human language processing and generation. 
Seq2Seq models use encoder-decoder architectures to map input sequences to output sequences, facilitating tasks like translation and synthesis. 
Then, this review summarizes the commonly used human emotion synthesis datasets from unimodal to cross-modal. 
Each dataset has annotated emotion labels according to its purpose, covering the simplest positive and negative emotion labels \cite{li2018delete} to complex 32 categories of compound emotion labels \cite{rashkin2019towards}.

\begin{figure}[t]
	\centerline{\includegraphics[width=1.0\linewidth]{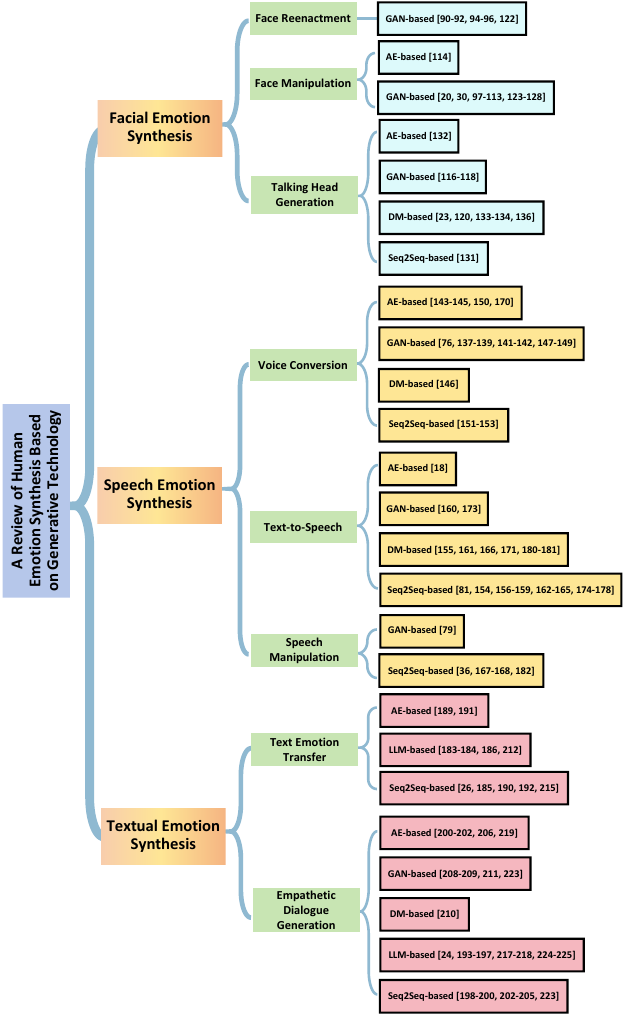}}
    \caption{Taxonomy of This Survey.}
    \label{fig:structure}
\end{figure}

Subsequently, we categorize human emotion synthesis into three subfields: facial emotion synthesis, speech emotion synthesis, and textual emotion synthesis. The taxonomy of this survey is shown in Fig.~\ref{fig:structure}.
Facial emotion synthesis involves modifying facial information in computer-generated faces to create more realistic and diverse emotions. 
This area of research intersects with face reenactment \cite{dhere2020review,luo2024codeswap}, face manipulation \cite{xie2023consistency}, and talking head generation \cite{zhen2023human, toshpulatov2023talking}.
Speech emotion synthesis involves altering the emotional attributes of speech segments or generating new speech that conveys specific emotions by manipulating acoustic features. 
This section will cover various studies on Voice Conversion \cite{sisman2020overview,zhou2022emotional}, Text-to-Speech (TTS) \cite{Kaur2023}, and speech manipulation tasks \cite{inoue2024fine}. 
Textual emotion synthesis refers to using computational techniques to infuse textual content with different emotions or sentiments, thereby enhancing its expressiveness. 
This section will focus on text emotion transfer and empathetic dialogue generation. 

Finally, this review summarize the prevalent evaluation metrics employed in the task, present key findings from multiple perspectives, and offer insights into future research directions based on the aforementioned systematic analysis. 
The main findings include:
(i) Generative models have made significant progress in emotion synthesis across multiple modalities, including facial images, speech, and text.
(ii) In the past, GAN-based methods have demonstrated strong capability in facial emotion synthesis, excelling at capturing subtle nuances in expressions. However, DMs have now emerged as a more promising alternative, offering superior control, stronger adaptability across different modalities.
(iii) Speech emotion synthesis has benefited from the adaptation of GANs and Seq2Seq models, with further improvements through AEs and DMs to enhance emotional depth and prosodic control.
(iv) Textual emotion synthesis has increasingly leveraged LLMs and Seq2Seq architectures, using sentiment control and emotional valence modulation to produce emotionally resonant content, although challenges remain in balancing emotional expressiveness with conversational coherence.
(v) Both subjective and objective metrics are essential for evaluating emotion synthesis models, with future research focusing on refining both to better capture emotional subtleties and align with human judgments.
\textcolor{blue}{
}
Looking ahead, there are still many directions worth exploring, which include:
(i) Combining different generative models like GANs, Seq2Seqs, AEs, DMs, and LLMs can enhance emotion synthesis by leveraging the strengths of each model for more accurate and realistic outputs.
(ii) Exploring new modalities like gestures, 
electroencephalogram (EEG) \cite{zheng2015investigating}, and electrocardiogram (ECG) \cite{geselowitz1989theory},
as well as cross-modal models, expanding the potential for immersive and interactive emotional experiences.
(iii) 
Real-time emotion generation on edge devices like smartphones and wearables can enable personalized, adaptive emotional interactions, with applications in healthcare, retail, and more.
(iv) 
Emotion synthesis can transform digital entertainment and filmmaking by enabling more authentic emotional expressions in virtual characters, enhancing storytelling, and allowing real-time emotional adjustments in films based on audience feedback.
This research provides a framework for understanding how generative models replicate human emotions, offering insights to guide future developments in the field.
Overall, the main contributions of this paper include:
\begin{itemize}
    \item To the best of our knowledge, this review provides the first systematic overview of human emotion synthesis based on generative technology.
    \item By analyzing more than 230 related papers, this review gives a taxonomy of generative technology-based human emotion synthesis in different modalities.
    \item 
    We summarize commonly used datasets and evaluation metrics for human emotion synthesis across different modalities.
    \item Finally, we discuss the current research status of human emotion synthesis based on generation technology and present a future outlook.
\end{itemize}

The remainder of this paper is structured as follows: Sections \ref{sec:Emotion Model} - \ref{sec:Databases} describe the gaps in existing review research, the mainstream emotion models, mathematical principles for generating models, and the commonly used datasets, Sections \ref{sec:Facial Emotion Synthesis} - \ref{sec:Textual Emotion Synthesis} introduce the latest human emotion synthesis works in the three modalities of face images, speech, and text, 
Section \ref{sec:Evaluation Metric} and \ref{sec:Discussion} summarize the common evaluation metrics in the field and discuss the current state of research and development trends, and Section \ref{sec:Conclusion} provides the conclusion. A list of abbreviations is given in Table \ref{Main acronyms}.

\begin{table}[H]
\centering
\caption{Main Acronyms.}
\label{Main acronyms}
\begin{tabular}{|p{1.2cm}|p{6cm}|}
\hline
Acronym & Full Form \\ \hline
AAE     & Adversarial Autoencoder \\ \hline
ACC     & Accuracy \\ \hline
AE      & Autoencoder \\ \hline
AIGC  & Artificial Intelligence Generated Content \\ \hline
AU     & Action Unit \\ \hline
AUD    & AU-intensity Discriminator \\ \hline
BLEU & Bilingual Evaluation Understudy  \\ \hline
CAAE     & Conditional Adversarial Autoencoder \\ \hline
CGAN    & Conditional GAN \\ \hline
CWT    &Continuous Wavelet Transform \\ \hline
DM      & Diffusion Model \\ \hline
FID   &Fréchet Inception Distance \\ \hline
F0 RMSE & F0 Root Mean Square Error \\ \hline
GAN     & Generative Adversarial Network \\ \hline
GPT     & Generative Pre-trained Transformer \\ \hline
MCD   &Mel Cepstral Distortion \\ \hline
MFCC    & Mel-Frequency Cepstral Coefficient \\ \hline
MOS  & Mean Opinion Score \\ \hline
PPL   & Perplexity \\ \hline
PSNR   &Peak Signal-to-Noise Ratio \\ \hline
RNN     & Recurrent Neural Network \\ \hline
Seq2Seq  &Sequence-to-Sequence \\ \hline
SMOS  & Similarity Mean Opinion Score  \\ \hline
SSIM    &Structural Similarity Index \\ \hline
TTS     & Text-to-Speech \\ \hline
T5      & Text-to-Text Transfer Transformer \\ \hline
VAE     & Variational Autoencoder \\ \hline
VC     & Voice Conversion \\ \hline
\end{tabular}
\end{table}

\begin{figure}[t]
	\centerline{\includegraphics[width=0.6\linewidth]{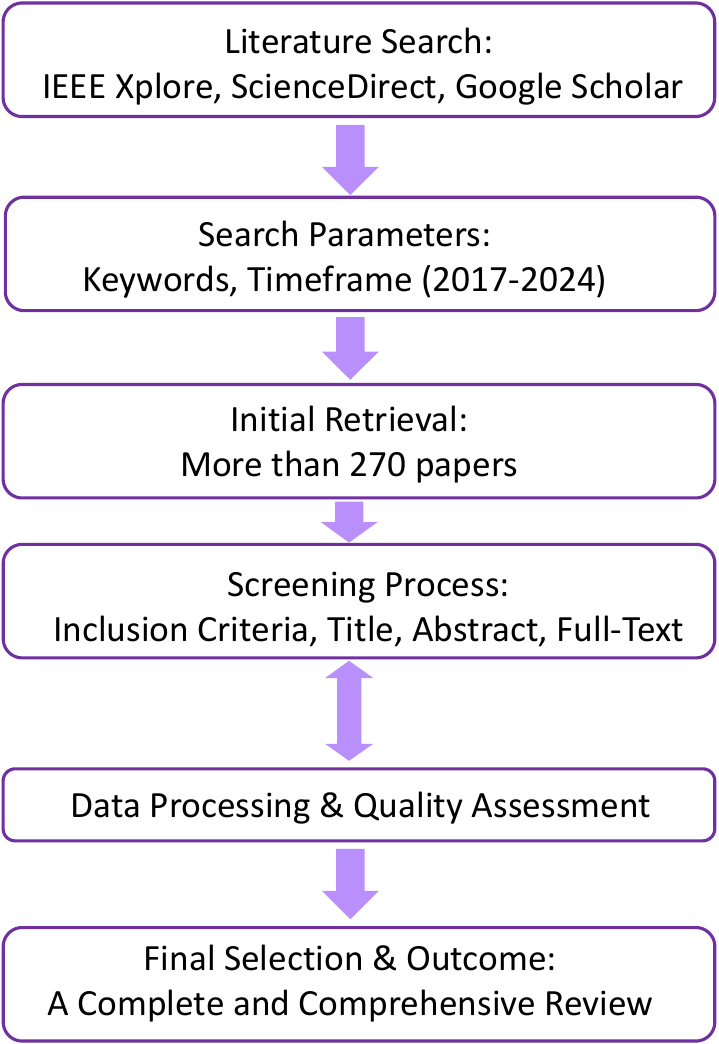}}
    \caption{A Comprehensive Review Methodology.} 
	\label{fig:Review Methodology}
\end{figure}

\section{Review Methodology}
\label{sec:Review Methodology}

To compile this extensive review on human emotion synthesis based on generative models, a systematic and rigorous approach was followed to ensure the comprehensiveness and relevance of the literature. The overall screening process is shown in Fig.~\ref{fig:Review Methodology}. Initially, we conducted comprehensive searches across key academic databases, including IEEE Xplore, ScienceDirect, and Google Scholar. The search strategy combined general and modality-specific keywords related to generative models and emotion synthesis. Examples of search terms included “facial emotion synthesis” + “generative models,” “speech emotion synthesis” + “generative models,” “emotional face reenactment,” and “emotional voice conversion,” etc., aiming to cover a wide array of studies in these domains. To ensure the inclusion of the most recent advancements, the search focused on papers published between 2017 and 2024, providing a comprehensive overview of developments in this timeframe. To further broaden the scope, we included terms referencing specific generative model architectures, such as AE, GAN, DM, LLM, and Seq2Seq. 
The initial search yielded more than 270 papers.

Furthermore, we established strict inclusion criteria for the selection process: (1) Peer-reviewed papers published up to November 2024, including both journal and conference papers; (2) Research focusing on the application of generative models to human emotion synthesis in various modalities; (3) 
Studies in which extensive experiments were conducted and fully evaluated.

After the initial retrieval, a two-step filtering process was applied to ensure the focus remained on emotion synthesis. Studies whose primary aim was not related to emotional generation or did not involve the use of generative models were excluded. 
Furthermore, we eliminated papers with incremental contributions to maintain the diversity of the sources reviewed. 
The final set of selected works provides a detailed overview of the current and impactful advancements in the field.

By following this structured methodology, we ensured the thoroughness and relevance of the selected studies, offering a timely and well-rounded perspective on the role of generative technology in human emotion synthesis across multiple modalities. 
This process ultimately enabled a focused analysis of both seminal works and the latest advancements, contributing to a deepened understanding of generative models in affective computing.

\section{Emotion Model}
\label{sec:Emotion Model}
Emotion is commonly understood as a complex and ever-changing state of mind and body, which can be triggered by various interactions\cite{brave2007emotion}, perceptions, or thoughts\cite{keltner1999functional}. It encompasses a wide range of experiences, cognitive evaluations, behavioral responses, physiological reactions \cite{mauss2010measures}, and communicative expressions. In the realm of human cognition, emotions play a crucial role in decision-making \cite{lerner2015emotion}, shaping our perceptions, and guiding our interactions with others \cite{parkinson2005emotion}.

\begin{figure}[t]
	\centerline{\includegraphics[width=1.0\linewidth]{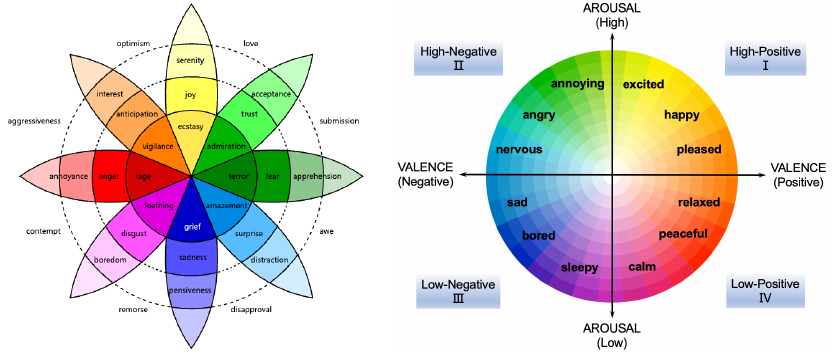}}
    \caption{Plutchik Wheel (left) and 2D Emotion Model (right).}
\scriptsize 
(1) Plutchik Wheel is sourced from 
\url{https://en.wikipedia.org/wiki/Robert_Plutchik#/media/File:Plutchik-wheel.svg}.\\ 
(2) Diagram of 2D emotion model is modified from \cite{khare2023emotion}.
 
	\label{fig:figure2}
\end{figure}

As illustrated in Fig.~\ref{fig:figure2}, the study of emotions has resulted in the development of different theoretical models, which can be primarily categorized into discrete emotions theory and multidimensional emotion theory \cite{khare2023emotion}. In the most basic discrete emotion framework, emotions are simply categorized as positive or negative, also known as polarity \cite{zhao2018personality,zhao2019personalized}. Within this framework, the term "emotion" is often replaced with "sentiment," which sometimes includes a neutral category as well. However, this sentiment categorization is considered too simplistic for certain contexts. Therefore, the more detailed discrete emotion theory categorizes basic emotions that are universally recognized across cultures into six or eight types \cite{ekman1992argument,plutchik2013theories,ma2022data}. On the other hand, the multidimensional emotions theory suggests that emotions can be viewed along a continuous spectrum, often defined by dimensions such as 2D (valence and arousal) \cite{wilson2003real} or 3D (valence, arousal, and dominance) \cite{mehrabian1996pleasure}. These theoretical perspectives provide valuable insights into the complex nature of human emotions and serve as foundational principles for emotion synthesis.

By considering emotions in generative tasks, machines not only understand and process information, but also become attuned to the emotional dimensions of human experience. This enriches the human-machine interaction landscape and opens up new avenues for the development of empathetic and intuitive AI developments.

\section{Difference}

\begin{table}[t]

    \caption{Differences between Our Work and Other Reviews.}
    \resizebox{\linewidth}{!}{%
    \begin{tabular}{|l|l|l|l|}
\hline
Reference &
  \begin{tabular}[c]{@{}l@{}}Related to\\generative models\end{tabular} &
  \begin{tabular}[c]{@{}l@{}}Related to \\ emotion synthesis\end{tabular} &
  \begin{tabular}[c]{@{}l@{}}Focus on \\ multiple modalities\end{tabular} \\ \hline
Our work                                                  & \multicolumn{1}{c|}{\checkmark\phantom{*}} &\multicolumn{1}{c|}{\checkmark\phantom{*}}& \multicolumn{1}{c|}{\checkmark\phantom{*}} \\ \hline
Kammoun et al. \cite{kammoun2022generative}               &  \multicolumn{1}{c|}{\checkmark*}          & \multicolumn{1}{c|}{\phantom{*}}          & \multicolumn{1}{c|}{\phantom{*}}          \\ \hline
Liu et al. \cite{liu2023gan}                              &  \multicolumn{1}{c|}{\checkmark*}          & \multicolumn{1}{c|}{\phantom{*}}          & \multicolumn{1}{c|}{\phantom{*}}          \\ \hline
Triantafyllopoulos et al. \cite{triantafyllopoulos2023overview} & \multicolumn{1}{c|}{\checkmark\phantom{*}} & \multicolumn{1}{c|}{\checkmark\phantom{*}} & \multicolumn{1}{c|}{\phantom{*}} \\ \hline
Wali et al. \cite{wali2022generative}                    & \multicolumn{1}{c|}{\checkmark*}           &  \multicolumn{1}{c|}{\checkmark\phantom{*}} & \multicolumn{1}{c|}{\phantom{*}}          \\ \hline
Zhang et al. \cite{zhang2023survey}                      & \multicolumn{1}{c|}{\checkmark\phantom{*}} & \multicolumn{1}{c|}{\checkmark\phantom{*}} & \multicolumn{1}{c|}{\phantom{*}}          \\ \hline
De et al. \cite{de2021survey}                            &  \multicolumn{1}{c|}{\checkmark*}          & \multicolumn{1}{c|}{\phantom{*}}          & \multicolumn{1}{c|}{\phantom{*}}          \\ \hline
Hajarolasvadi et al. \cite{hajarolasvadi2020generative}  &\multicolumn{1}{c|}{\checkmark*}           & \multicolumn{1}{c|}{\checkmark\phantom{*}} & \multicolumn{1}{c|}{\checkmark\phantom{*}} \\ \hline
\end{tabular}
}
\raggedright
\scriptsize *: The article only relates to GAN.
	\label{table2}
\end{table}

The current state of research in affective computing primarily focuses on areas such as human sentiment analysis \cite{poria2017review}, emotion detection, and emotion recognition \cite{ma2024generative}. These tasks have a long-standing tradition and are considered significant when viewed in a broader context. For instance, Saxena et al. \cite{saxena2020emotion} conducted a survey on emotion recognition methods, examining facial, physiological, speech, and text approaches. They highlighted key techniques like Stationary Wavelet Transform and Particle Swarm Optimization. In another study, Nandwani et al. \cite{nandwani2021review} analyzed sentiment analysis and emotion detection methods, discussing the transition from lexicon-based to deep learning techniques. They addressed challenges and emphasized the need for advanced and versatile models to improve accuracy and adaptability across different domains and languages. Canal et al. \cite{canal2022survey} presented a systematic literature review on facial emotion recognition from images. They categorized the techniques into classical and neural network-based approaches, highlighting the slightly higher accuracy of classical methods compared to neural networks, despite the latter's generalization capabilities.

Generative models currently constitute one of the mainstream directions in artificial intelligence research.
However, most of the existing reviews only focus on the synthesis of a single modality, such as facial images \cite{kammoun2022generative,liu2023gan}, speech \cite{triantafyllopoulos2023overview,wali2022generative}, and text \cite{zhang2023survey,de2021survey}, and many of them ignore the emotional aspects of the synthesis process. 
The most relevant work to ours is \cite{hajarolasvadi2020generative}. 
This paper provides a thorough review of GANs in synthesizing human emotions, with a focus on facial expressions, speech, and cross-modal synthesis. It details various GAN architectures, their applications in emotion synthesis, challenges faced, and future directions. By evaluating numerous studies, it highlights how GANs enhance emotion recognition accuracy, offer data augmentation, and create realistic, diverse emotional samples across modalities. 
However, the key distinction between our survey and the aforementioned one lies in the fact that ours exclusively focuses on emotion synthesis. 
In addition, this review introduces a series of generation models other than GANs, such as the  Seq2Seq model in the field of emotional speech synthesis, and LLMs in text emotion synthesis.

In summary, the differences are shown in Table~\ref{table2}. Our survey is based on a clear definition of human emotion synthesis, focusing on the application of generative models in this emerging field, as well as the latest research developments in various sub-fields.

\section{Generative Model}
The generative model \cite{oussidi2018deep} refers to a model that can be described as generating data, belonging to a type of probability model. 
In machine learning, it can model data directly or establish conditional probability distributions between variables via Bayes’ theorem, allowing the generation of new data not present in the training set.
In this section, 
we explain the mathematics of different generative models, including AE, GAN, DM, LLM, and Seq2Seq.

\subsection{Auto-Encoder (AE)}
AEs \cite{hinton2006reducing} are neural network models used in generative tasks to efficiently learn data representations.
These models compress input data into simplified patterns, then reconstruct it by preserving key features while minimizing errors in the reproduction process.
As a variant of AE, Variational Auto-Encoders (VAEs) introduced by Kingma and Welling in 2013 \cite{kingma2013auto}, offer a principled approach to learning latent data representations to account for data uncertainty and variability, making them well-suited for generative tasks. 
The training of VAEs is guided by the maximization of the Evidence Lower BOund (ELBO), which can be expressed as follows:
\begin{equation}
\text{ELBO} = \mathbb{E}_{q_\phi(z|x)}[\log p_\theta(x|z)] - D_{KL}[q_\phi(z|x) \| p(z)],
\label{elbo}
\end{equation}
where the encoder, \(q_\phi(z|x)\), maps the input data \(x\) to a distribution over the latent space characterized by parameters \(\phi\). Typically, this distribution is assumed to be Gaussian, with the encoder outputting the mean and variance of the distribution. The latent variable \(z\), sampled from this distribution, is then fed into the decoder, \(p_\theta(x|z)\), which attempts to reconstruct the input data, where \(\theta\) denotes the parameters of the decoder. 
In the Equation~(\ref{elbo}), the first term is the expected log-likelihood of the data given the latent variables, encouraging accurate reconstruction of the data. 
The second component measures how closely the encoded data patterns match an expected statistical distribution \(p(z)\), using the Kullback-Leibler divergence formula.
By optimizing the ELBO, VAEs learn to balance the trade-off between fidelity in data reconstruction and adherence to a structured latent space.

\subsection{Generative Adversarial Network (GAN)}

GANs \cite{goodfellow2014generative} are distinguished by their unique training methodology, which leverages the concept of adversarial learning, setting up a dynamic competition between two distinct neural networks: the generator and the discriminator. The generator network, \(G\), aims to map latent space vectors, drawn from a prior distribution \(p_z(z)\), to data space, effectively generating new data samples that mimic the distribution of real data, \(p_{data}(x)\). 
In contrast, the discriminator network, \(D\), is trained to distinguish between samples drawn from the real data distribution and those produced by the generator. 
The competition between networks steadily improves their performance, ultimately enabling the generator to create lifelike outputs.
The training of GANs is formulated as a min-max game, which can be formally represented by the following value function \(V(G, D)\):
\begin{equation}
\begin{aligned}
\min_G \max_D V(D, G) = &\mathbb{E}_{x \sim p_{data}(x)}[\log D(x)] +\\ &\mathbb{E}_{z \sim p_z(z)}[\log(1 - D(G(z)))],
\end{aligned}
\end{equation}
where the first term represents the expected log-probability that the discriminator correctly identifies real data samples as real. The second term represents the expected log-probability that the discriminator correctly identifies fake samples (generated by \(G\)) as fake. Training GANs involves alternating between optimizing \(D\) to maximize \(V(D, G)\) for fixed \(G\) (improving \(D\)'s accuracy in distinguishing real from fake samples) and optimizing \(G\) to minimize \(V(D, G)\) for fixed \(D\) (improving \(G\)'s ability to generate realistic samples). This adversarial training process continues until a state of equilibrium is reached where \(G\) generates samples indistinguishable from real data by \(D\). The adversarial training mechanism of GANs has proven to be highly effective for generating complex, high-dimensional data.

\subsection{Diffusion Model (DM)}
DMs \cite{ho2020denoising} have emerged as a class of powerful generative models that synthesize data by gradually refining random noise into structured patterns. The fundamental principle behind DMs involves two phases: a forward (noising) phase and a reverse (denoising) phase. 
In the forward phase, step by step, the model gradually adds random noise to the data until the original information becomes completely obscured.
This process is described by a Markov chain that gradually corrupts the original data distribution, \(x_0\), into a tractable noise distribution, \(x_T\), over \(T\) steps. Mathematically, this can be expressed through a sequence of conditional probabilities:
\begin{equation}
q(x_t | x_{t-1}) = \mathcal{N}(x_t; \sqrt{1-\beta_t} x_{t-1}, \beta_t I),
\end{equation}
where \(\beta_t\) represents the variance of the noise added at each step, and \(I\) is the identity matrix. The sequence of \(\beta_t\) values is predefined to control the noise level at each step, ensuring a smooth transition from data to noise.
The reverse phase aims to learn the reverse process, modeling the conditional distribution of \(x_{t-1}\) given \(x_t\), effectively denoising the data. The model, typically parameterized by a neural network, is trained to approximate the reverse conditional probabilities:
\begin{equation}
p_\theta(x_{t-1}|x_t) = \mathcal{N}(x_{t-1}; \mu_\theta(x_t, t), \Sigma_\theta(x_t, t)),
\end{equation}
where the parameters \(\mu_\theta\) and \(\Sigma_\theta\) are functions learned during training, with \(\theta\) representing the model's parameters. 
The model learns by minimizing the gap between real and generated data distributions, using either statistical bounds or direct probability optimization.
Despite their computational intensity, DMs have emerged as a cornerstone in generative modeling due to their impressive performance in generating high-quality samples.

\subsection{Large Language Model (LLM)}
Large Language Models are AI systems trained on massive text datasets, using billions of parameters to learn language patterns. Their deep understanding of language enables human-like text comprehension and generation, transforming how computers process natural language.
The advent of Transformer architecture, developed by Vaswani et al. in 2017 \cite{vaswani2017attention}, marked a significant breakthrough in language modeling.

The training objective of LLMs can generally be described as learning a conditional probability distribution over sequences. Given an input sequence \(X = (x_1, x_2, \dots, x_T)\), the model maximizes the likelihood of generating each token based on prior tokens, represented as:
\begin{equation}
p(X) = \prod_{t=1}^{T} p(x_t | x_{<t})
\end{equation}
where \(p(x_t | x_{<t})\) is the probability of token \(x_t\) given its context \(x_{<t}\). A key feature of this process is the attention mechanism, which allows the model to dynamically focus on different parts of the context at each step. In self-attention, the probability of generating each token is influenced by a weighted sum of the context, with attention weights \(\alpha_{tj}\) computed as:
\begin{equation}
\alpha_{tj} = \frac{\exp(\frac{q_t \cdot k_j}{\sqrt{d_k}})}{\sum_{k=1}^{T} \exp(\frac{q_t \cdot k_k}{\sqrt{d_k}})}
\end{equation}
and the new token representation \(z_t\) is given by:
\begin{equation}
z_t = \sum_{j=1}^{T} \alpha_{tj} v_j
\end{equation}
By analyzing relationships between tokens, the attention system helps the model understand context and produce coherent, relevant text.

So far, several LLMs have gained prominence, including Generative Pre-trained Transformer (GPT) \cite{radford2018improving} series, Bidirectional Encoder Representations from Transformers (BERT) \cite{devlin2019bert}, eXtreme Language Understanding Network (XLNet) \cite{yang2019xlnet}, and Text-to-Text Transfer Transformer (T5) \cite{raffel2020exploring}, etc. These models have been pre-trained on vast amounts of text data and can effectively fine-tune for specific tasks, such as language translation, sentiment analysis, and text generation \cite{radford2019language,liu2024mllm}.

\subsection{Sequence-to-Sequence (Seq2Seq) Model}  
Seq2Seq model \cite{sutskever2014sequence} is a neural network architecture designed for tasks involving sequential input-output pairs.
It follows an encoder-decoder structure, where each component is typically implemented with recurrent neural networks (RNNs) or Transformers in later models.

One defining feature of Seq2Seq models is their focus on sequential data, which differentiates them from models like GANs or VAEs that do not inherently account for sequential dependencies. 
Unlike general language models, Seq2Seq models specialize in transforming one sequence into another, effectively capturing both immediate and distant patterns in the data.

In the Seq2Seq model, the encoder processes the input sequence \((x_1, x_2, \dots, x_T)\) to produce a context vector \(c\), often represented by the encoder’s final hidden state:
\begin{equation}
c = h_T
\end{equation}
where \(h_T\) encapsulates the input sequence's essential information. The decoder then uses this context vector to generate the output sequence \((y_1, y_2, \dots, y_{T'})\) one element at a time, conditioned on \(c\) and previously generated outputs:
\begin{equation}
s_t = g(y_{t-1}, s_{t-1}, c)
\end{equation}
\begin{equation}
p(y_t | y_{<t}, x) = \text{softmax}(Ws_t)
\end{equation}
where \(s_t\) is the hidden state at time \(t\), and \(W\) is a weight matrix for calculating the output distribution.

Seq2Seq models are effective for handling variable-length input and output sequences, making them well-suited for applications like translation, summarization, and question answering, where coherent sequence transformation is required. The differences between Seq2Seq models and other generative models are shown in Table~\ref{relationseq2seq}.

\begin{table}[ht]
\centering
\caption{Relationship Between Seq2Seq and Other Generative Models.}
\begin{tabular}{|>{\centering\arraybackslash}m{1.7cm}|>{\centering\arraybackslash}m{1.7cm}|>{\arraybackslash}m{4cm}|} 
\hline
\textbf{Model Type} & \textbf{Belongs to Seq2Seq?} & \textbf{Description} \\ \hline
LLMs & Partially belongs  & Some LLMs (e.g., T5, BART) are implementations of Seq2Seq, but LLMs cover a broader range. \\ \hline
AEs & Does not belong & Similar to Seq2Seq in architecture, but with different goals and tasks. \\ \hline
GANs & Does not belong & Completely different model type with unrelated goals and structures. \\ \hline
DMs & Does not belong & Entirely different generative models, unrelated to Seq2Seq. \\ \hline
\end{tabular}
\label{relationseq2seq}
\end{table}

\section{Databases}
\label{sec:Databases}
The performance of human emotion synthesis tasks based on generative models is closely tied to the quality and richness of the utilized datasets. 
To be specific, 
the diversity and scope of the datasets play a crucial role in the model's ability to generalize across various emotional states, cultural contexts, and individual differences. 
The structure and content of emotion databases directly shape how researchers design and build emotion synthesis models.
The structure, annotation scheme, and inherent biases of the datasets influence the choice of model architecture, loss functions, and training strategies. 
Furthermore, the size and quality of the datasets influence the choice between end-to-end learning approaches and modular architectures, with large, high-quality datasets enabling end-to-end learning of emotion synthesis, while smaller or noisier datasets might necessitate the use of pre-trained components or transfer learning techniques. 
Based on these emotional datasets of different modalities, such as facial images, speech, and text, the designed models can imitate human emotional expressions with high precision from different aspects.
Table~\ref{table3} summarizes the common datasets used in the field of human emotion synthesis, providing a comprehensive overview of the available resources for researchers and practitioners in this domain.

\begin{table*}[t]
    \centering
    \caption{Common Datasets for Human Emotion Synthesis with Generative Models.}
    \resizebox{\linewidth}{!}{%
    \begin{tabular}{|l|l|l|l|l|l|l|}
\hline
Database &
  Year&
  Modalities &
  Samples &
  Subjects &
  Category \\ \hline
Oulu-CASIA \cite{zhao2011facial} &2011
  &visual
   &480 sequences
   &80
   &happiness, surprise, sadness, anger, fear, disgust
  \\ \hline
RaFD \cite{langner2010presentation} &2010
  &visual
   &8040 images
   &49
   &sad, neutral, angry, contemptuous, disgusted, surprised, fearful, happy
  \\ \hline
CK+ \cite{lucey2010extended}&2010
   &visual
   &593 images 
   &123
   &happiness, sadness, surprise, fear, anger, disgust, neutral, contempt
  \\ \hline
CFEE \cite{du2014compound}&2014
  &visual
   &229 images
   &230
   &happiness, surprise, sadness, anger, fear, disgust
  \\ \hline
AffectNet \cite{mollahosseini2017affectnet}&2017 
  &visual
   &450,000 images 
   &/
   &happiness, sadness, surprise, fear, anger, disgust, neutral, contempt
  \\ \hline
DISFA \cite{mavadati2013disfa}&2013
  &visual
   &130,788 images
   &27
   &continuous annotation of graded changes in spontaneous facial expression of emotion
  \\ \hline
EmotioNet \cite{fabian2016emotionet}&2016
  &visual
   &1,000,000 images
   &/
   &23 basic or compound emotion categories(happy, sad, fearful, angrily surprised, sadly angry, etc.)
  \\ \hline

ESD \cite{zhou2021seen} &2021
  &audio
   &350 utterances
   &20
   &happy, sad, neutral, angry, surprise
  \\ \hline
EmoV-DB \cite{adigwe2018emotional}&2018
  &audio
   &7590 utterances
   &5
   &neutral, amused, angry, sleepy, disgust
  \\ \hline
Emo-DB \cite{burkhardt2005database}&2005
  &audio
   &800 sentences
   &10
   &neutral, anger, fear, joy, sadness, disgust, boredom 
  \\ \hline
MEmoSD \cite{jia2019gan}&2019
  &audio
   &/
   &4
   &angry, happy, neutral, sad 
  \\ \hline
CaFE \cite{gournay2018canadian}&2018
  &audio
   &936 samples
   &12
   &neutral, sadness, happiness, anger, fear, disgust, surprise 
  \\ \hline
KES \cite{im2022emoq}&2019
  &audio
   &21,000 speeches
   &1
   &neutral, happy, sad, angry, surprised, fearful, disgusted 
  \\ \hline
ETOD \cite{im2022emoq}&2019
  &audio
   &6000 speeches
   &13
   &neutral, happy, sad, angry 
  \\ \hline
YELP review \cite{li2018delete}&/
  &text
   &6,990,280 reviews
   &/
   &positive, negative
  \\ \hline
AMAZON review \cite{he2016ups}&/
  &text
   &/
   &/
   &positive, negative
  \\ \hline
EmpatheticDialogue \cite{rashkin2019towards} &2019
  &text
   &24,850 conversations 
   &810
   &32 emotion labels (suprised, excited, angry, proud, sad, annoyed, grateful, etc.)
  \\ \hline
MojiTalk \cite{zhou2018mojitalk}&2018
  &text
   &662,159 conversations
   &/
   &64 emoji labels
  \\ \hline
  
MEAD \cite{wang2020mead}&2020
  &visual + audio
   &281,400 clips
   &60
   &angry, disgust, contempt, fear, happy, sad, surprise, neutral
  \\ \hline
CREMA-D \cite{cao2014crema}&2014 
  &visual + audio
   &7442 utterances
   &91
   &happiness, surprise, sadness, anger, fear, disgust
  \\ \hline
EmoVoxCeleb \cite{albanie2018emotion}&2018 
  &visual + audio
   &153,500 tracks
   &1251
   &neutral, happiness, surprise, sadness, anger, disgust, fear, contempt
  \\ \hline
SAVEE \cite{haq2008audio}&2008
  &visual + audio
   &480 utterances
   &4
   &neutral, anger, disgust, fear, happiness, sadness, surprise 
  \\ \hline 
  
  RAVDESS \cite{livingstone2018ryerson}&2018 
  &visual + audio
   &7356 videos
   &24
   &Happiness, Sadness, Surprise, Fear, Anger, Disgust, Neutral, Contempt
  \\ \hline
  
IEMOCAP \cite{busso2008iemocap}&2008 
  &visual +  audio + text
   &10,039 samples
   &10
   &categorical and continuous annotations
  \\ \hline

\end{tabular}%
    }
	\label{table3}
\end{table*}

\section{Facial Emotion Synthesis}
\label{sec:Facial Emotion Synthesis}
Facial emotion synthesis is a crucial research field within human emotion synthesis, aiming to generate faces that express specified emotions. This technology holds significant academic value in computer graphics and computer vision, 
while also demonstrating great promise for applications in virtual reality (VR), gaming, and interactive computer systems.
Based on existing works, we can broadly categorize facial emotion synthesis into three main approaches: face reenactment (Section \ref{sec:Face reenactment}), talking head generation (Section \ref{sec:Talking head generation}), and facial manipulation (Section \ref{sec:Facial Manipulation}).
The related works are illustrated in Table \ref{face table}.

\subsection{Face Reenactment}
\label{sec:Face reenactment}
Face reenactment focuses on transferring facial expressions from a source actor to a target face, preserving the identity of the target while adopting the emotional expressions of the source.  
This technique is particularly useful in applications like film dubbing, virtual avatars, and privacy-preserving video conferencing.

\begin{figure}[h]
  \centerline{\includegraphics[width=1.0\linewidth]{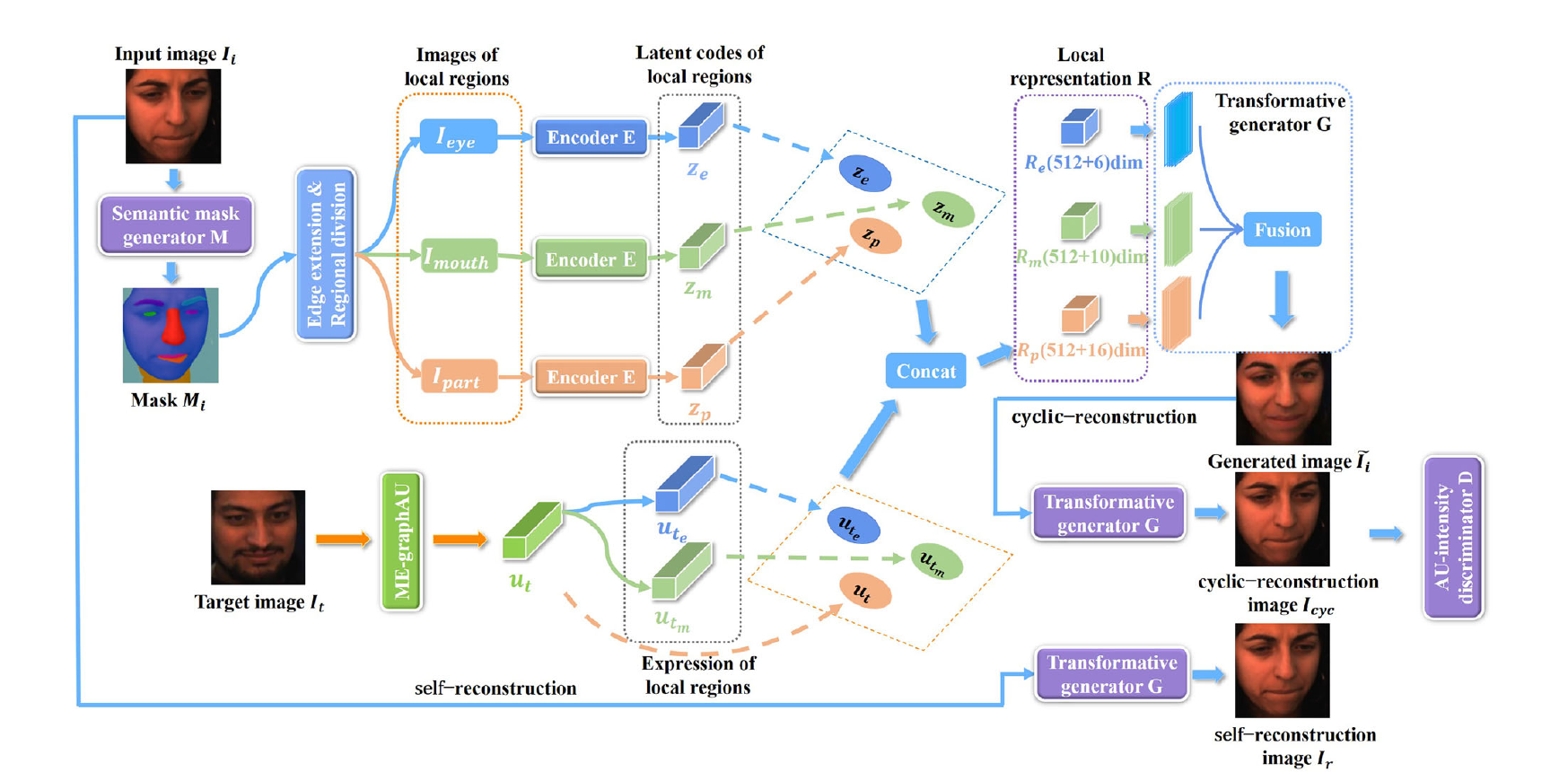}}
     \label{transfer}
  \caption{A mask-based GAN \cite{xue2024semantic} for face reenactment. The system included four main components. A Semantic Mask Generator (SMG) produced masks for specific facial regions (eyes, mouth, cheeks). Then these masks were encoded into latent codes through an Adversarial Autoencoder (AAE). A Transformative Generator (TG) used these codes along with target expression labels to generate new facial expressions, with an AU-intensity Discriminator (AUD) that assessed their quality and intensity. }
  \label{fig:transfer}
\end{figure}

In facial reenactment, there are some tasks that emphasize emotional attributes in the face. 
For example, 
in \cite{tripathy2020icface}, Tripathy et al. introduced ICface, a GAN-based face animator that manipulated facial expressions in a given image. 
The animation process was guided by interpretable control signals, such as head pose angles and Action Units (AU) values, which were derived from various sources, allowing for selective emotion transfer. 
Zeng et al. \cite{zeng2020realistic} proposed DAE-GAN, which employed two deforming autoencoders to separate identity and pose in unlabeled videos, reducing the need for manual annotation. 
It realized emotional transfer between different identities with varied poses using conditional generation and disentangled features. 
Strizhkova et al. \cite{strizhkova2021emotion} proposed a novel method for emotion editing in head reenactment videos by manipulating the latent space of a pre-trained GAN.
This technique disentangled emotion, identity, and pose within the latent space, allowing for the direct modification of emotions in the reenactment videos without affecting the person's identity or the speech-related facial expressions. 
Groth et al. \cite{groth2020altering} designed a new method to achieve emotion mapping by generating correctly recognized expressions, using video reenactments to influence the intensity of the emotion.
In \cite{ali2019all}, Ali et al. utilized two encoders to separately capture expression from a source and identity from a target image, merging these features to create expressive images, enhanced by innovative consistency losses for both expression and identity features. In Fig. \ref{fig:transfer}, Xue et al. presented LSGAN \cite{xue2024semantic}, employing a transformative generator that combined target expression labels with specific facial region features to produce clear and distinct facial expressions in images. 
Shao et al. \cite{shao2021wp2} utilized dual parallel generators and wavelet-based discriminators for facial expression translation, enhancing realism by focusing on key areas with an attention mechanism and capturing expression details across scales without the bidirectional translation interference seen in single-generator models.

\subsection{Face Manipulation}
\label{sec:Facial Manipulation}
Face manipulation involves editing specific attributes of a face, such as changing expressions, age, hairstyles, etc., to generate different versions of the same person. 
It can be seen that this has a completely different goal compared to face reenactment which transfers the expressions and movements of a source face to a target face.
Researchers in this field focus on the alteration of specific facial attributes while preserving the remaining attributes unchanged, under the condition of explicit predefined facial information, thereby effecting changes in emotional expression. 
Works in this field are mainly based on GAN and its variants \cite{doukas2021headgan, zhao2021action, xia2021local, wang2019dft, akram2023sargan, zhang2021gated, akram2024us, azari2024emostyle, xie2023consistency, Apolito2021ganmut, peng2019apprgan, pumarola2018ganimation}.
For instance, 
in \cite{xie2023consistency}, Xie et al. proposed a novel approach, using GAN equipped with consistency preservation and feature entropy regularization techniques. This innovation achieved improved results in attribute translation, particularly in preserving consistency and reducing feature entropy. 
Song et al. \cite{song2018geometry} utilized facial geometry to guide expression creation but required a neutral face image, making the process more complex. It consisted of two GANs for changing and removing expressions.
In \cite{kong2021dualpathgan}, Kong te al. proposed a dual-path GAN for emotion synthesis and introduced a learning strategy based on a separation discriminator to train it more efficiently.
In \cite{tang2020fine}, Tang et al. introduced structured latent space and perceptual loss to achieve fine-grained expression manipulation while preserving identity and global facial shape.
Patashnik et al. \cite{patashnik2021styleclip} innovatively utilized text-driven manipulation techniques, in conjunction with the extraordinary visual concept encoding abilities of CLIP and StyleGAN, to specify desired attributes like different emotional facial expressions.
Liu et al. \cite{liu2022evogan} first proposed a novel general framework, EvoGAN, that combined an evolutionary algorithm (EA) and GAN to work as a whole, which generated face images with more compound expressions. 
In \cite{tang2019expression}, Tang et al. proposed a novel ECGAN for generating faces with different emotions based on the input expression attribute vector. 
And in \cite{sola2023unmasking}, the innovation in Sola et al.'s work was the use of ECGAN, a novel expression-conditioned GAN, to incorporate expression information into unmasking processes, resulting in improved naturalness and expression fidelity of generated faces. 
In \cite{lindt2019facial}, Lindt et al. updated the Conditional Adversarial Autoencoder (CAAE) \cite{makhzani2015adversarial} framework to manipulate facial expressions in images based on continuous two-dimensional emotion labels, representing valence and arousal.

\begin{figure}[h]
  \centerline{\includegraphics[width=1.0\linewidth]{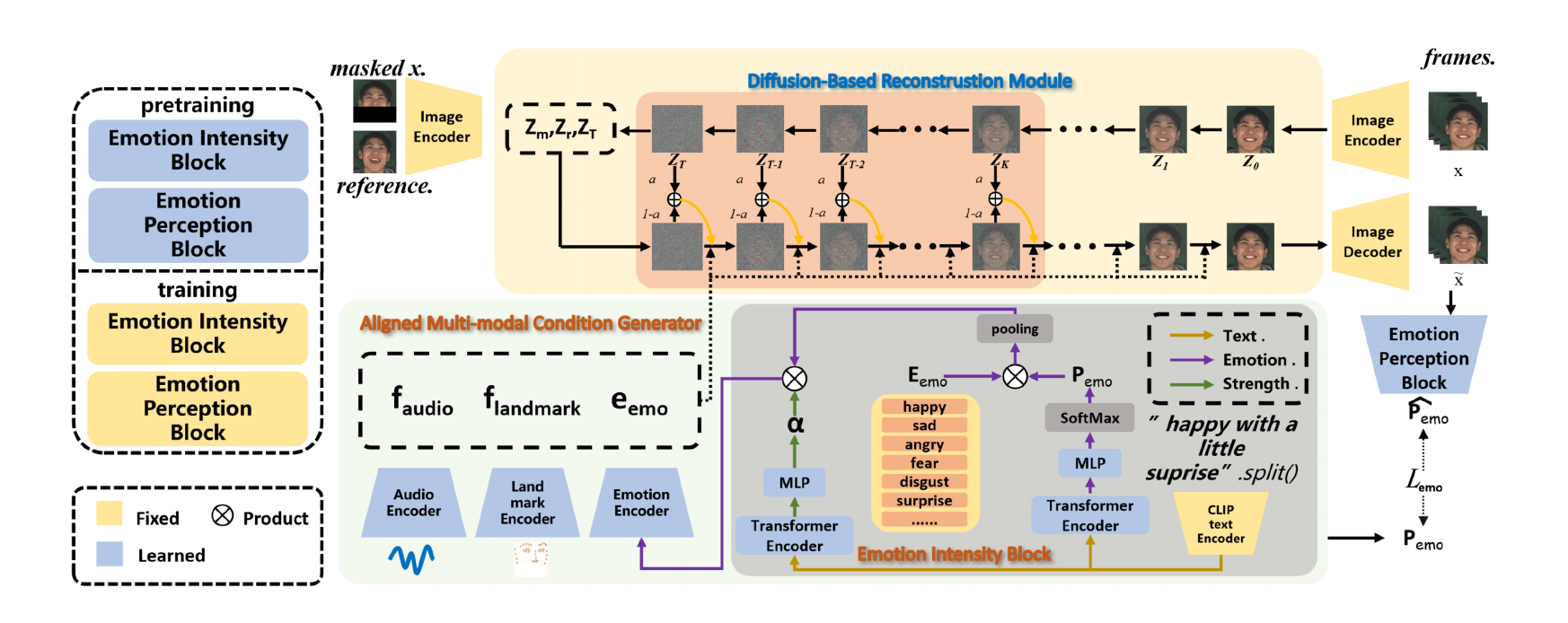}}
  \caption{A DM-based talking head generation model from \cite{zhang2024emotalker}. This architecture started by extracting conditions from multiple modalities: audio, visual, and textual inputs. Specifically, the emotion intensity block used a pretrained CLIP text encoder to convert text prompts into embeddings that represented the underlying emotional content. These embeddings, reflecting nuanced emotional states and their strengths, were then integrated into the DM's denoising process.}
  \label{fig:face editing}
\end{figure}

\subsection{Talking Head Generation}
\label{sec:Talking head generation}
Talking head generation aims to create realistic, animated facial models that can speak and emote based on input audio or text.  
This approach is particularly valuable in creating virtual assistants, digital newscasters, and personalized content for educational or entertainment purposes.

A series of works that incorporate emotion synthesis into talking head generation have been explored.
For example,
Eskimez et al. \cite{eskimez2021speech} presented a novel system for generating talking faces, achieving independent control of emotional expressions by disregarding the emotions expressed in the input speech audio and instead conditioning the face generation on an independent categorical emotion variable. 
In \cite{vougioukas2020realistic}, Vougioukas et al. 
developed a specialized GAN that uses three discriminators to create detailed facial expressions that match a speaker's emotional state.
As illustrated in Fig.~\ref{fig:face editing}, Zhang et al. presented EmoTalker \cite{zhang2024emotalker}, a framework for emotionally editable talking face generation, utilizing a diffusion model and a novel Emotion Intensity Block, and integrating a custom dataset to enhance emotion interpretation. 
In \cite{zeng2020talking}, Zeng et al. presented ET-GAN, an end-to-end system for generating talking faces with tailored expressions from guiding videos, identity images, and arbitrary audio, utilizing multiple encoders for identity, expression, and audio-lip synchronization, alongside advanced frame and spatial-temporal discriminators. 
Gan et al. \cite{gan2023efficient} generated synchronized emotion coefficients and emotion-driven facial images through flow-based and vector-quantized models, while using lightweight emotion prompts or CLIP supervision for model control and adaptation. 
In \cite{tan2024flowvqtalker}, Tan et al. used the regularized flow model to generate emotion-related expression coefficients, mapped the input emotion coefficients to the latent space, and generated diverse expression dynamics based on audio and emotion labels, while achieving high-fidelity restoration of emotional details through a vector quantization image generator. 
In another work \cite{tan2024edtalk}, they used orthogonal basis decomposition to decompose facial dynamics into independent latent spaces of lip shape, head posture, and emotional expression, allowing the model to adjust the relevant expression coefficients in each independent latent space. 

\clearpage

\begin{table*}[t]
\centering
\caption{Literature on Generative Models for Face Reenactment, Face Manipulation, and Talking Head Generation, in Facial Emotion Synthesis.}
\rotatebox{90}{  
    \begin{adjustbox}{max width=\textheight}
\begin{tabular}{|l|l|l|l|l|l|}
\hline
Reference                         & Year     & Model     &Dataset &Performance      \\ \hline

                     Ali et al. \cite{ali2019all}                & 2019 & TER-GAN      &Oulu-CASIA& User study   \\ \hline
                     Shao et al. \cite{shao2021wp2}              & 2021 & WP2-GAN   &RaFD/CFEE&Accuracy: 89.47, 87.97/FID: 41.74, 24.91/SSIM: 0.6818, 0.6659       \\ \hline  
                     Tripathy et al. \cite{tripathy2020icface}   & 2020  &GAN  & VoxCeleb& Image Quality Assesment scores: 25.02, 33.08           \\ \hline 
                     Zeng et al. \cite{zeng2020realistic}        & 2020 &DAE-GAN  &VoxCeleb1/RaFD&SSIM: 0.65, 0.73/FID: 60.8, 13.8/User Study: 0.61  \\ \hline 
                     Strizhkova et al. \cite{strizhkova2021emotion}    &2021    &GAN   &MUG/MEAD &ECS: 0.88, 0.58/FID: 19.7, 25.5/ACD: 0.10, 0.13        \\ \hline
                     Groth et al. \cite{groth2020altering}         &2020    &Encoder-Decoder   &MoCap &Ratings figure for perceived intensity and sincerity        \\ \hline
                     Xue et al. \cite{xue2024semantic}           & 2024 &LSGAN   &RaFD/DISFA  &FID: 31.121/PSNR: 23.417/SSIM: 0.858        \\ \hline
                     Hu et al. \cite{hu20232cet}                 & 2023 &2CET-GAN &CFEE/RaFD &FID: 31.1/IS: 1.55/Expression Transfer Score: 3.37  \\ \hline

                    Choi et al. \cite{choi2018stargan}       & 2018    &StarGAN   & CelebA/RaFD& Classification error: 2.12\%       \\ \hline 
                    Liu et al. \cite{liu2022evogan}       & 2022 &EvoGAN        & EmotioNet/RaFD  & User study    \\ \hline 
                    Xie et al. \cite{xie2023consistency}   & 2023 &GAN    &FFHQ/RaFD/LFW&FID: 8.32-16.05/Accuracy: 97.42-98.61     \\ \hline 
                    Tang et al. \cite{tang2020fine} &  2020&EGGAN    &RaFD/MMI&MSE: 0.24/PCC: 0.66         \\ \hline 
                    Patashnik et al. \cite{patashnik2021styleclip}       & 2021   & StyleGAN     &FFHQ & User study        \\ \hline 
                    Sola et al. \cite{sola2023unmasking}            & 2023  & ECGAN      &RAFDB &Accuracy: 78.54/SSIM: 0.52-0.64/FID: 244.75-668.68/Perceptual Loss: 1.21-1.46      \\ \hline
                    Zhu et al. \cite{zhu2019ugan}      &  2019 &UGAN   &CFEE  &FID: 44.8        \\ \hline  
                    Ding et al. \cite{ding2018exprgan} &  2018&ExprGAN     &Oulu-CASIA& User study         \\ \hline  
                    Tesei et al. \cite{tesei2019generating}    & 2019   &CCycleGAN    &FER2013& FID figure       \\ \hline 
                    Tang et al. \cite{tang2019expression}      & 2019&ECGAN     &AR/Yale/JAFFE/FERG/3DFE   &AMT Score: 35.32 VGG Score: 78.13, 80.32    \\ \hline 
                    Wang et al. \cite{wang2019comp}        & 2019   &Comp-GAN  &CelebA/$F^2\text{ED}$  &Accuracy figure          \\ \hline
                    Lindt et al. \cite{lindt2019facial}          & 2019 &CAAE    &AffectNet &RMSE: 0.528, 0.607/SAGR: 0.567, 0.483/CCC: 0.312, 0.210       \\ \hline
                     Kong et al. \cite{kong2021dualpathgan}        & 2021  &DualPathGAN     &EmoVoxCeleb/VoxCeleb2&PSNR: 32.56/SSIM: 0.953/FID: 9.20        \\ \hline
                     Doukas et al. \cite{doukas2021headgan}        & 2021  &HeadGAN     &VoxCeleb&FID: 50.9/FCD: 334/CSIM: 0.716        \\ \hline 
                     Zhao et al. \cite{zhao2021action}        & 2021  &PAttGAN &DISFA/DISFACat &ICC: 0.89, 0.88/MAE: 0.24, 0.28/MSE: 0.22, 0.29/FID: 27.32, 21.77             \\ \hline
                     Xia et al. \cite{xia2021local}        & 2021  & LGP-GAN &RaFD &IS: 1.31/FID: 11.88  \\ \hline              
                     Wang et al. \cite{wang2019dft}        & 2019 & GAN &EmotioNet/RaFD  &User study           \\ \hline   
                     Akram et al. \cite{akram2023sargan}       & 2023  &SARGAN &KDEF/CFEE/RaFD &ACD: 0.306/FVS: 94.18±1.11/FID: 53.95    \\ \hline 
                     Zhang et al.  \cite{zhang2021gated}       & 2021  & SwitchGAN &RaFD &FID: 17.4/Accuracy: 98.32      \\ \hline 
                     Akram et al.   \cite{akram2024us}        & 2024 &US-GAN &KDEF/RaFD/CFEE  &ACD: 0.2832/FVS: 94.19±1.11/User study: 43\% improvement(expression plausibility)      \\ \hline     
                     Azari et al.   \cite{azari2024emostyle}       & 2024  &StyleGAN &CelebA/FFHQ &LPIPS: 0.07/FID: 7.86/ID: 0.88       \\ \hline      
                     Apolito et al.   \cite{Apolito2021ganmut}      & 2021  &GAN &AffectNet &Smoothness score:0.33-0.38/ERE: 0.020, 0.018/FED: 0.71              \\ \hline   
                     Peng et al.   \cite{peng2019apprgan}       & 2019  &ApprGAN &Bosphorus/CK+/MUG &Correlation coefficients: 0.972, 0.953, 0.975/Normalised distances: 2.061, 1.879, 1.835             \\ \hline 
                     Pumarola et al. \cite{pumarola2018ganimation}     & 2018    & GAN&EmotioNet/RaFD &User study            \\ \hline 
                     Ding et al. \cite{ding2018exprgan}          & 2018 &ExprGAN            &Oulu-CASIA& User study   \\ \hline
                     Song et al. \cite{song2018geometry}         & 2018   &G2-GAN   &CK+/Oulu-CASIA &    User study     \\ \hline
                     Xu et al. \cite{xu2024self} &2024 &StyleGAN &MEAD/RAVDESS &Realism: 0.40/Emotion similarity: 0.36/Mouth shape similarity: 0.43 \\ \hline
                    Eskimez et al. \cite{eskimez2021speech}       & 2021  &GAN     &CREMA-D &PSNR: 30.91/SSIM: 0.85/Accuracy: 55.3       \\ \hline
                     Zhang et al. \cite{zhang2024emotalker}          & 2024 &Diffusion model     &MEAD/CREMA-D/FED &CSIMD: 0.67, 0.51/Accuracy: 84.76, 75.13       \\ \hline
                     Vougioukas et al. \cite{vougioukas2020realistic}        & 2020   &GAN  &CREMA-D  &PSNR: 23.565/SSIM: 0.70/ACD: $1.40 \cdot 10^{-4}$       \\ \hline
                     Tan et al. \cite{tan2023emmn}         & 2023 &Encoder-Decoder   &CFD/MEAD/CREMA-D &Accuracy: 65.20/PSNR: 29.38, 30.03/SSIM: 0.66, 0.68/M-LMD: 2.78, 3.03/F-LMD: 2.87, 3.16      \\ \hline
                     Tan et al. \cite{tan2024style2talker}      & 2024   &Encoder-Decoder &MEAD &SSIM: 0.795/FID: 23.207/M-LMD: 3.317/F-LMD: 2.696            \\ \hline
                     Zhai et al. \cite{zhai2023talking}      & 2023   &Seq2Seq &MEAD/CREMA-D &SSIM: 0.82, 0.79/PSNR: 30.29, 30.55/M-LMD: 2.14, 1.16/LMD: 2.44, 1.46/FID: 19.59, 42.53/EP: 80.52, 85.64              \\ \hline
                     Sheng et al. \cite{sheng2023stochastic}    & 2023      &DVAE&MEAD/CREMA-D&SSIM: 0.79, 0.92/LPIPS: 0.07, 0.08/LMD: 1.83, 1.34/LVD: 1.56, 0.84/CSIM: 0.88, 0.87/Emoacc: 0.87, 0.84             \\ \hline
                     Gan et al. \cite{gan2023efficient}       & 2023 &Encoder-Decoder &VoxCeleb2/MEAD&PSNR: 21.75/SSIM: 0.68/FID: 19.69/M-LMD: 2.25/F-LMD: 2.47/Accuracy: 75.43            \\ \hline
                     Tan et al. \cite{tan2024flowvqtalker}     & 2024 &Diffusion model &HDTF/MEAD&SSIM: 0.708, 0.689/FID: 15.165, 16.553/M-LMD: 1.643, 1.939/F-LMD: 1.958, 2.061/Accuracy: 71.53           \\ \hline
                    Tan et al. \cite{tan2024edtalk}     & 2024 &Encoder-Decoder &HDTF/MEAD &PSNR: 26.504, 22.771/SSIM: 0.845, 0.769/FID: 13.172, 15.548/M-LMD: 1.197, 1.102/F-LMD: 1.111, 1.060/Accuracy: 68.85           \\ \hline
                     Ma et al. \cite{ma2023dreamtalk}         & 2023 &Diffusion model   &MEAD/HDTF/Voxceleb2&SSIM: 0.86, 0.85, 0.69/CPBD: 0.16, 0.31, 0.30/F-LMD: 1.93, 1.80, 2.69/M-LMD: 2.91, 2.15, 2.72          \\ \hline
                     Zhang et al. \cite{zhang2023dream}        & 2023    &Diffusion model   &MEAD/HDTF&LPIPS: 0.169, 0.176/CPBD: 0.299, 0.280/F-LMD: 3.845, 3.948        \\ \hline
                     Zeng et al. \cite{zeng2020talking}          & 2020  &ET-GAN   &CREMA-D/GRID&PSNR: 23.981, 25.771/SSIM: 0.733, 0.810/FID: 76.92, 61.33       \\ \hline
                     Sun et al. \cite{sun2023continuously}  &2023 &StyleGAN &MEAD/RAVDESS &LIE: 0.739/CPBD: 0.247, 0.166/FED: 9.40/FID: 30.36/ID: 0.931/LSE-C: 7.11/LSE-D: 8.03 \\ \hline
                     Wang et al. \cite{wang2024eatface} &2024 &Diffusion model &MEAD/CREMA-D &PSNR: 32.6131, 34.3401/CPBD: 0.3803, 0.5180/Emo-Acc: 74.57 \\ \hline
\end{tabular}%
\label{face table}
\end{adjustbox}
}
\end{table*}

\clearpage

\section{Speech Emotion Synthesis}
Speech emotion synthesis is a critical research field within human emotion synthesis, focusing on the manipulation and generation of acoustic features in speech signals to alter or create specific emotional states. 
This area of study aims to modify key vocal parameters such as volume, intonation, pitch, speaking rate, and timbre to effectively convey a desired emotional state in synthesized speech.
Based on existing works, we divide it into voice conversion (Section \ref{sec:Voice Conversion}) ,text-to-speech (Section \ref{sec:Text-to-Speech}) and speech manipulation (Section \ref{sec:Speech Manipulation}).
Table \ref{speech table} illustrates the overall papers about speech emotion synthesis.

\subsection{Voice Conversion}
\label{sec:Voice Conversion}
Voice conversion technology modifies a speaker's voice to express different emotions while keeping the original words intact. This enables emotional dubbing in films and games, and helps create more natural-sounding AI assistants.

\begin{figure}[h]
  \centerline{\includegraphics[width=1.0\linewidth]{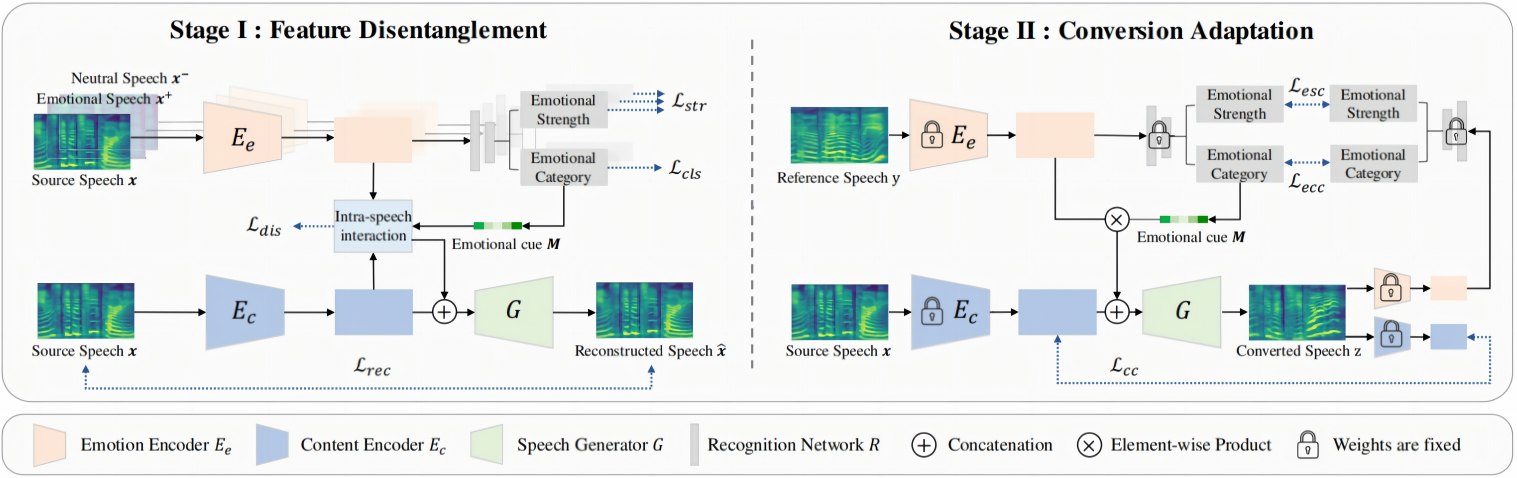}}
  \caption{ A two-stage model \cite{chen2023attention} for voice conversion. In the first stage, the method separated emotional and content features from speech using an attention-based mechanism, incorporating inter-speech relation to model emotional strength. The second stage employed a conversion adaptation strategy, leveraging a multi-view consistency mechanism to ensure that the emotional nuances were accurately transformed while the core speech content remained intact. This framework facilitated precise control over the emotional output of synthesized speech.   }
\label{fig: speech editing}
\end{figure}

For instance, in \cite{zhou2021seen}, Zhou et al. leveraged VAW-GAN and deep emotional features from speech emotion recognition to describe emotional prosody. 
By using adjustable emotional features to guide the decoder, this approach could transform speech to express both familiar and new emotions.
In \cite{zhou2020converting}, they combined VAW-GAN and Continuous Wavelet Transform (CWT) for advanced spectrum and prosody conversion. By incorporating CWT-based F0 modeling, the system uniquely enhanced the granularity of prosody representation. 
In \cite{zhou2021vaw}, they focused on disentangling and re-composing emotional elements in speech, innovatively employing CWT for detailed prosody analysis and integrating F0 conditioning in the decoder to enhance emotion conversion performance. 
Similarly, in \cite{zhou2020transforming}, they proposed a CycleGAN-based model \cite{zhu2017unpaired} that did not require parallel data. 
It utilized CWT to analyze F0 on multiple scales, enabling detailed prosody modification. 
In \cite{fu2022improved}, Fu et al. also presented an improved CycleGAN-based model that incorporated a transformer to augment temporal dependencies and integrated curriculum learning and a fine-grained level discriminator, enhancing the model's ability to capture and convert emotional nuances in speech more effectively. 
Rizos et al. \cite{rizos2020stargan} utilized class-conditional GAN and an auxiliary domain classifier to generate emotional speech samples. 
In \cite{elgaar2020multi}, Elgaar et al. introduced a Factorized Hierarchical Variational Autoencoder (FHVAE) for multi-speaker and multi-domain emotional voice conversion, enhancing disentangled representation and emotion conversion quality through novel algorithms and loss functions. 
Gao et al. \cite{gao2018nonparallel} used style transfer autoencoders for emotional voice conversion without parallel data, leveraging disentangled representation learning to modify emotion-related characteristics. 
In Fig. \ref{fig: speech editing}, Chen et al. introduced a Tacotron2-based framework using emotion disentangling modules \cite{chen2023attention} to achieve cross-speaker emotion transfer by separating speaker identity from emotion.
In \cite{oh2024durflex}, Oh et al. proposed the DurFlex-EVC model, integrating a style autoencoder and unit aligner for advanced control and flexibility. It leveraged HuBERT features, denoising diffusion models, and self-supervised learning to modify emotional tones while maintaining linguistic content and unique vocal traits. 
Moreover, there were also some works related to emotional voice conversion based on StarGAN \cite{du2021expressive, meftah2023english, shah2023nonparallel}, AE \cite{du2021disentanglement}, and Seq2Seq \cite{zhou2022emotion, kreuk2022textless, choi2021sequence}.

\subsection{Text-to-Speech (TTS)}
\label{sec:Text-to-Speech}
TTS aims to generate natural-sounding speech based on the semantic content and context of the input text. This technology is particularly crucial for enabling more engaging and human-like interactions with virtual assistants, audiobook narrations, and personalized content delivery. Some researchers have attempted to incorporate emotional factors into TTS systems to synthesize speech with emotional expressiveness.

\begin{figure}[h]
  \centerline{\includegraphics[width=1.0\linewidth]{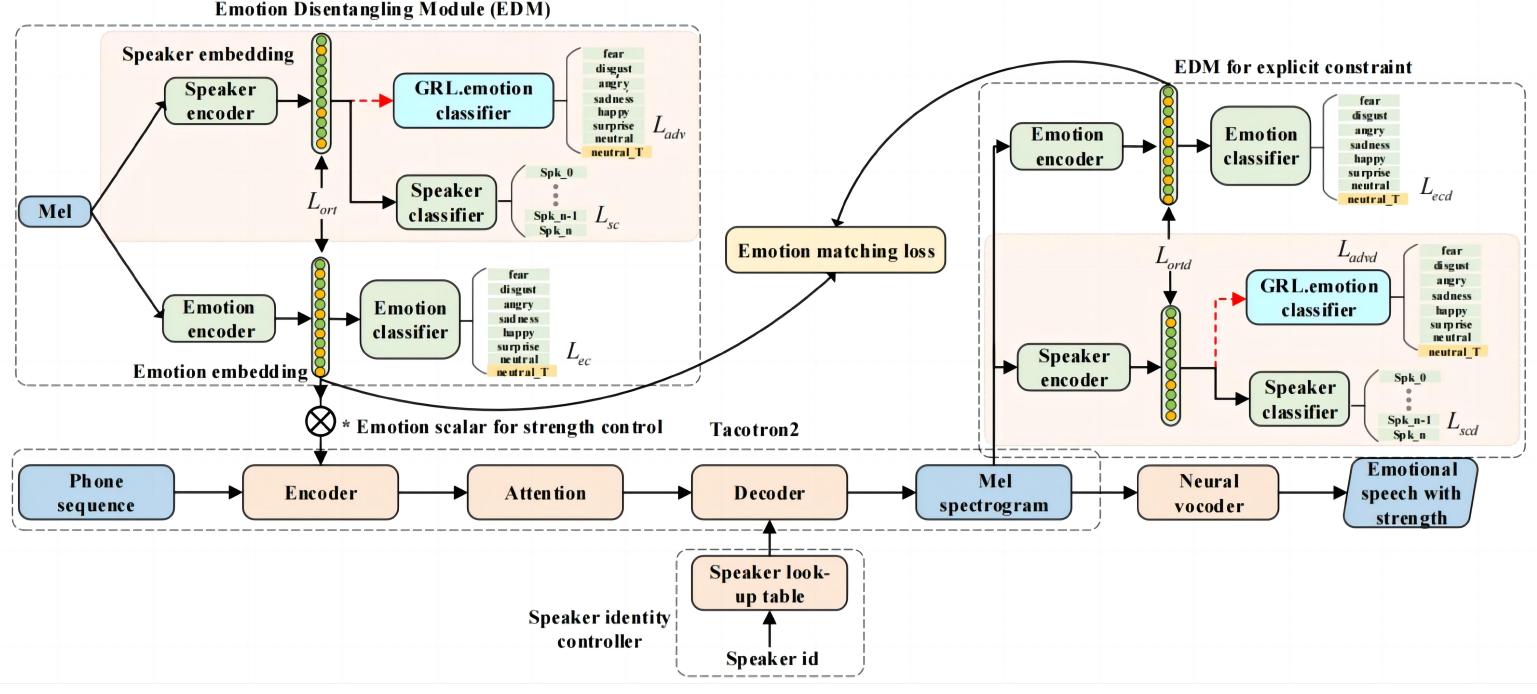}}
  \caption{A Tacotron2-based TTS model from \cite{li2022cross}. This system integrated an Emotion Disentangling Module (EDM) that utilized dual encoders to separate emotion-related features from speaker identity, refining the emotion embedding for clearer expression without speaker leakage. An identity controller maintained the target speaker's identity by incorporating speaker-specific information, enabling the synthesis of emotionally expressive speech consistent with the target speaker’s vocal characteristics. }
  \label{fig: speech generation}
\end{figure}

For example, in \cite{lei2021fine}, Lei et al. introduced a unified model for fine-grained emotional speech synthesis, obtaining fine-grained emotion expressions with emotion descriptors or phoneme-level manual labels. 
Li et al. \cite{li2023diclet} developed DiCLET-TTS, which improves emotional speech synthesis by combining emotion separation techniques with advanced probability-based decoding.
In \cite{tits2020exploring}, Tits et al. explored adapting a Deep Convolutional TTS (DCTTS) model to various emotions using minimal emotional data. 
In \cite{schnell2021improving}, Schnell et al. leveraged WaveNet and emotion intensity extraction using attention LSTM and transformer models, demonstrating increased perceived emotion accuracy. 
Wu et al \cite{wu2019end} synthesized emotional speech with limited labeled data, achieving comparable performance to fully supervised models. 
In \cite{um2020emotional}, Um et al. employed an inter-to-intra distance ratio algorithm and an effective interpolation technique to achieve nuanced emotion intensity control. 
In \cite{im2022emoq}, Im et al. proposed EmoQ-TTS, a system that synthesized expressive emotional speech by conditioning phoneme-wise emotion information with fine-grained emotion intensity, using intensity pseudo-labels generated via distance-based intensity quantization. 
Hortal et al. \cite{hortal2021gantron} combined Tacotron 2 with GANs to modulate prosody, allowing customization of inferred speech with specified emotions. 
Guo et al. \cite{guo2023emodiff} presented "EmoDiff," a DM-based model that enabled intensity-controllable emotional speech synthesis using a soft-label guidance technique.
Li et al. \cite{li2021controllable} utilized the Tacotron framework, enhanced with emotion classifiers and style loss, to generate expressive, controllable emotional speech efficiently.
In \cite{lei2022msemotts}, Lei et al. introduced a novel method integrating global-level, utterance-level, and local-level modules to achieve precise emotion modeling and transfer, allowing for versatile emotional expression in synthesized speech. 
In Fig. \ref{fig: speech generation}, Li et al. 
created a system \cite{li2022cross}  that extracts emotion patterns independent of the speaker's voice, allowing emotions to be transferred between speakers and adjusted in intensity.
In \cite{li2021towards}, Li et al. 
developed a technique that analyzes speech style at different scales, capturing both broad and fine details to produce more controllable and expressive synthetic voices.
In \cite{tang2023emomix}, Tang et al. introduced mix methods, enabling the manual combination of noise at runtime to produce diverse emotional mixtures, which was validated through evaluations demonstrating its capability to generate speech with various mixed emotions.

\subsection{Speech Manipulation}
\label{sec:Speech Manipulation}

Speech manipulation focuses on modifying various aspects of speech signals, such as the speaker's identity or linguistic content. It shares similarities with face manipulation in Section \ref{sec:Facial Manipulation}, as both aim to alter or control specific attributes of human-generated data, be it facial features or speech characteristics. 
Recently, some researchers have begun exploring how to manipulate the emotional attributes of speech.

For instance, in \cite{jia2019gan}, Jia et al. presented ET-GAN, an innovative cross-language emotion transfer system for speech synthesis, which was unique for not necessitating parallel training data and utilized CycleGAN, achieving significant improvements in emotional accuracy and naturalness of synthetic speech. 
In \cite{matsumoto2020controlling}, Matsumoto et al. utilized WaveNet and auxiliary features like voiced, unvoiced, and silence flags to generate speech-like emotional sounds without linguistic content, enhancing emotional expressiveness control.
In \cite{wang2024emotion}, Wang et al. presented Emo-CampNet, a text-based speech editing model. It integrated emotion attributes and a context-aware mask prediction network, employing generative adversarial training and data augmentation to enhance emotional expressiveness and speaker variability in edited speech. 
In \cite{inoue2024fine}, Inoue et al. introduced a novel emotion editing technique in speech synthesis utilizing a hierarchical emotion distribution extractor within the FastSpeech2 framework \cite{ren2019fastspeech}, enabling fine-grained, quantitative emotion control at phoneme, word, and utterance levels for dynamic and nuanced speech generation.

\clearpage

\begin{table*}[t]
\centering
\caption{Literature on Generative Models for Voice Conversion, Text-to-Speech and Speech Manipulation, in Speech Emotion Synthesis.}
\rotatebox{90}{  
    \begin{adjustbox}{max width=\textheight}
\begin{tabular}{|l|l|l|l|l|}
\hline
 Reference                            & Year   & Model      &Dataset &Performance      \\ \hline

                    Zhou et al. \cite{zhou2021seen}      & 2021   &VAW-GAN    &ESD/IEMOCAP &MCD: 4.127-4.916/MOS: 3.24, 2.94, 3.15/Preference test     \\ \hline
                    Fu et al. \cite{fu2022improved}      & 2022   &CycleGAN   &Japanese emotional speech dataset/ESD    &MOS figures/MCD: 5.65, 16.38/F0 RMSE: 44.91, 105.99       \\ \hline 
                    Zhou et al. \cite{zhou2021vaw}    & 2021 &VAW-GAN  &EmoV-DB  &MCD: 4.085, 4.278/LSD: 5.681, 6.106/F0 RMSE: 61.712, 52.348/PCC: 0.893, 0.867/Preference test        \\ \hline
                    Zhou et al. \cite{zhou2020transforming} &  2020&CycleGAN     &Emotional speech corpus &MCD: 10.23, 8.71/F0 RMSE: 65.05, 63.03/PCC: 0.76, 0.76/Preference test        \\ \hline
                    Rizos et al. \cite{rizos2020stargan}         & 2020 &StarGAN   &IEMOCAP &Spectrogram representation/Human evalution score          \\ \hline 
                    Zhou et al. \cite{zhou2020converting}         & 2020     &VAW-GAN   &EmoV-DB/English emotional speech corpus/JL-Corpus &MCD: 4.439, 4.683/LSD: 6.161, 6.275/PCC: 0.776, 0.691/MOS: 2.808   \\ \hline 
                    Gao et al. \cite{gao2018nonparallel} &  2018 &AE   &IEMOCAP &Emotion conversion MOS: 48\%/Speaker similarity MOS: 3.55         \\ \hline 
                    Zhou et al. \cite{zhou2022emotion} &  2022&Seq2Seq &VCTK/ESD &MOS figures/MCD: 4.13, 4.15, 4.25/DDUR: 0.24, 0.17, 0.31       \\ \hline 
                    Oh et al. \cite{oh2024durflex}  &  2024&AE Diffusion model    &ESD &nMOS: 3.70/sMOS: 3.63/eMOC: 72.97/UTMOS: 3.58/ECA: 91.58/SECS: 74.83         \\ \hline 
                    Elgaar et al.  \cite{elgaar2020multi} &  2020&VAE   &Collected dataset/IEMOCAP  &Subjective evalution figures           \\ \hline  
                    Kreuk et al.  \cite{kreuk2022textless} &  2022 &Seq2Seq &VCTK/EmoV &MOS figures/eMOC figures             \\ \hline  
                    Du et al.  \cite{du2021disentanglement}&  2021 &VAE   &ESD &MCD figures/F0 RMSE figures/SV accuracy figures/MOS: 3.53, 3.58, 3.74/Preference test          \\ \hline 
                    Du et al.  \cite{du2021expressive}     &  2021 &StarGAN  &ESD &MCD figures/MOS: 3.044/Preference test           \\ \hline  
                    Chen et al.  \cite{chen2023attention}  &  2023 &AE   &ESD &MCD: 4.596/ACC: 0.830/RMSE: 0.117/Emotion similarity: 76.12\%          \\ \hline  
                    Meftah et al.  \cite{meftah2023english}  &  2023&StarGAN &ESD &MCD,F0 RMSE figures/Waveforms figures/Confusion matrices/Spectrograms  figures        \\ \hline  
                    Shah et al.  \cite{shah2023nonparallel}  &  2023 &StarGAN   &Hindi Emotional Speech Database &MOS: 4.21±0.01/Similarity: 0.76/EmoAcc: 41.50/Preference test          \\ \hline   
                    Choi et al.  \cite{choi2021sequence}     &  2021 &Seq2Seq   &Plain-to-emotional dataset &MOS: 4.14/MCD: 5.778, 16.055/Emotion confusion matrices/Emotion similarity figures          \\ \hline   
                    Qi et al. \cite{qi2024realisticemotional}        & 2024  &CVAE   &ESD &MCD: 4.06/RMSE: 38.28/DDUR: 0.21/MSD: 19.61, 20.55, 21.54, 22.24     \\ \hline

                   
                    Zhang et al. \cite{zhang2023iemotts}   & 2023 &AE   &Multi-S60-E3/Child-S1-E6/VA-S2/Read-S40 &MOS: 4.09±0.03, 3.76±0.05, 4.10±0.03 (Emotion Similarity,Speaker Similarity,Voice Quality)       \\ \hline  
                    Lei et al. \cite{lei2021fine}      & 2021        &Seq2Seq    &Internal corpus&MCD: 4.91, 5.03/F0 trajectories        \\ \hline 
                    Li et al. \cite{li2023diclet}      & 2023        &Diffusion model  & 5 female monolingual speakers&Emotion similarity DMOS: 3.90-4.04/Cosine Similarity: 0.25, 0.73              \\ \hline 
                    Tits et al.\cite{tits2020exploring}   & 2020     &Seq2Seq  &EmoV-DB&MOS: 2.00, 2.10, 2.27, 3.59, 3.29 (amused, angry, disgusted, neutral, sleepy)          \\ \hline 
                    Schnell et al. \cite{schnell2021improving}&  2021 &Seq2Seq  &SAVEE/IEMOCAP/WSJCAM0&Accuracy: 35.5, 28.9(total, emo)           \\ \hline 
                    Wu et al. \cite{wu2019end}       & 2019           &Seq2Seq   &Emotional speech corpus  &MCD: 2.64/F0 RMSE: 64.4/V/UV: 8.26/FFE: 21.81          \\ \hline  
                    Um et al. \cite{um2020emotional}    & 2020        &Seq2Seq   &Korean male voice database  &MOS: 3.90±0.54/Recognition acurracy/Preference test       \\ \hline  
                    Im et al. \cite{im2022emoq}       &  2022&Seq2Seq  &KES/ETOD  &MOS: 3.72, 3.95/MCD: 4.81, 2.94/F0 RMSE: 53.15, 30.61/EmoAcc: 99.39, 86.85          \\ \hline  
                    Hortal et al. \cite{hortal2021gantron}  &  2021   &Seq2Seq GAN     &LJ Speech/VESUS/CREMA-D/RAVDESS  &Accuracy         \\ \hline 
                    Guo et al. \cite{guo2023emodiff}    & 2023        &Diffusion model    &ESD &MOS: 4.13±0.10/MCD: 5.94/Classification accuracy/Preference test         \\ \hline
                    Kang et al. \cite{kang2023zet}    & 2023           &Diffusion model  &Multi-emotional dataset/ESD/LibriTTS Test &ECA: 51.59, 38.89, 39.86, 32.57/MOS: 3.44, 3.31/SMOS: 3.22           \\ \hline  
                    Zhu et al. \cite{zhu2019controlling}     & 2019   &Seq2Seq    &Emotional speech corpus&  Mel spectrograms/Pitch/PCA ordination diagram trajectories       \\ \hline
                    Tang et al. \cite{tang2023emomix}   & 2023        &Diffusion model  &IEMOCAP/ESD &MOS: 4.10, 3.92/SMOS: 4.02, 3.82/MCD: 5.29, 5.65       \\ \hline  
                    Lei et al. \cite{lei2022cross}        & 2022       &GAN &Internal corpus &Emotion MOS: 4.04, 4.01, 3.86/Pearson Correlation: 0.776, 0.790, 0.759, 0.794/F0 curves            \\ \hline
                    Lei et al. \cite{lei2022msemotts}         & 2022   &Seq2Seq &Emotional speech corpus &MCD: 3.63/MOS: 4.02±0.119/CMOS: 0.520, 0.342/Preference: 58.1, 51.2/F0 curves         \\ \hline 
                    Li et al. \cite{li2021controllable}      & 2021    &Seq2Seq  &Emotional speech corpus &Peference test/Strength confusion matrices/pitch trajectories            \\ \hline
                    Li et al. \cite{li2022cross}       & 2022          &Seq2Seq    &DB\_1/AIC/DB\_6 &Emotion similarity DMOS: 3.71±0.066/Cosine simlarity: 0.28, 0.60        \\ \hline
                    Li et al. \cite{li2021towards}        & 2021        &Seq2Seq   &Internal corpus &MOS: 4.155±0.552, 4.136±0.701          \\ \hline
                    Cai et al. \cite{cai2021emotion}      & 2021       &Seq2Seq  &IEMOCAP/BC2013-English/RECOLA   &MOS: 3.66 Accuracy: 78.75, 91.0          \\ \hline
                    Wang et al. \cite{wang2023fine}        & 2023      &Seq2Seq      &EmoV-DB  &MCD: 4.66/MOS: 3.76±0.03/Accuracy: 0.67, 0.67, 0.77/preference test     \\ \hline
                    Zhou et al. \cite{zhou2022speech}      & 2022       &Seq2Seq  &VCTK/ESD &MCD figures/PCC figures/MOS: 3.21-3.81/BWS test            \\ \hline
                    Diatlova et al. \cite{diatlova2023emospeech}        & 2023  &Seq2Seq   &ESD &MOS: 4.37/NISQA: 4.1        \\ \hline
                    Guan et al. \cite{guan2024mmtts}        & 2024  &Seq2Seq   &MEAD-TTS &MOS: 4.36, 3.95, 4.11, 3.77, 4.19, 3.93/SMOS: 4.61, 4.03 /MCD: 3.17, 6.69/EmoAcc: 0.659, 0.261, 0.636, 0.318      \\ \hline
                    Zhou et al. \cite{zhou2024emotionaldimension}        & 2024  &Encoder-Decoder   &ESD/LibriTTS &MOS: 3.67/CMOS: 0.64, 0.71, 0.15, 0.92/preference test      \\ \hline
                    Tang et al. \cite{tang2024edtts}        & 2024  &Diffusion model   &IEMOCAP &MOS: 4.12/SMOS: 4.10/ERA: 0.749/EDER: 27.8     \\ \hline
                    Jing et al. \cite{jing2024enhancing}        & 2024  &Diffusion model  & ESD/MSP-Podcast &MOS-Q: 3.88/MOS-S: 3.36, 3.22, 3.05, 3.34/preference test      \\ \hline
                    Jia et al. \cite{jia2019gan}      & 2019 &ET-GAN  &IEMOCAP/Emo-DB/CaFE/MEmoSD &FAD,naturalness MOS figures/Preference test         \\ \hline  
                    Inoue et al. \cite{inoue2024fine}        & 2024  &Seq2Seq &Blizzard/ESD &MOS: 3.596±0.141/MCD: 4.348/Pitch Distortion: 1.151/Energy Distortion: 4.018/FD: 6.881/BWS test           \\ \hline
                    Wang et al. \cite{wang2024emotion}        & 2024   &CampNet   &VCTK/ESD &F0 curve/MCD: 3.078, 3.495, 3.528, 3.425, 3.332/Preference test/MOS figures        \\ \hline  
                    Matsumoto et al. \cite{matsumoto2020controlling}        & 2020  &WaveNet   &JSUT corpus&MOS figures/F0 distribution/Confusion matrices (Subject-perceived emotions)        \\ \hline
                    Shi et al. \cite{shi2024rset}        & 2024  &Encoder-Decoder   &ESD &MOS: 3.98/SMOS: 4.15/EmoAcc: 99.31/MCD: 4.81        \\ \hline

\end{tabular}%
\label{speech table}
\end{adjustbox}
}
\end{table*}

\clearpage

\section{Textual Emotion Synthesis}
\label{sec:Textual Emotion Synthesis}
Textual emotion synthesis is a vital research field within human emotion synthesis, focusing on the generation of texts that possess specific emotional, sentiment, or empathetic attributes. 
This area of study aims to create or modify written content to convey desired emotional states, sentiments, or empathetic responses, thereby enhancing the expressiveness and impact of textual communication.
Based on existing works, we 
consider the following two types of tasks:
text emotion transfer (Section \ref{sec:Text Emotion Transfer}) and empathetic dialogue generation (Section \ref{sec:Empathetic Dialogue Generation}).
Table \ref{text table} shows the literature about textual emotion synthesis.

\subsection{Text Emotion Transfer}
\label{sec:Text Emotion Transfer}
Text emotion transfer focuses on transforming the emotional tone or sentiment of an existing text while preserving its core semantic content. It allows for the modification of neutral text into emotionally charged content, or the alteration of one emotional state to another.

\begin{figure}[]
  \centerline{\includegraphics[width=1.0\linewidth]{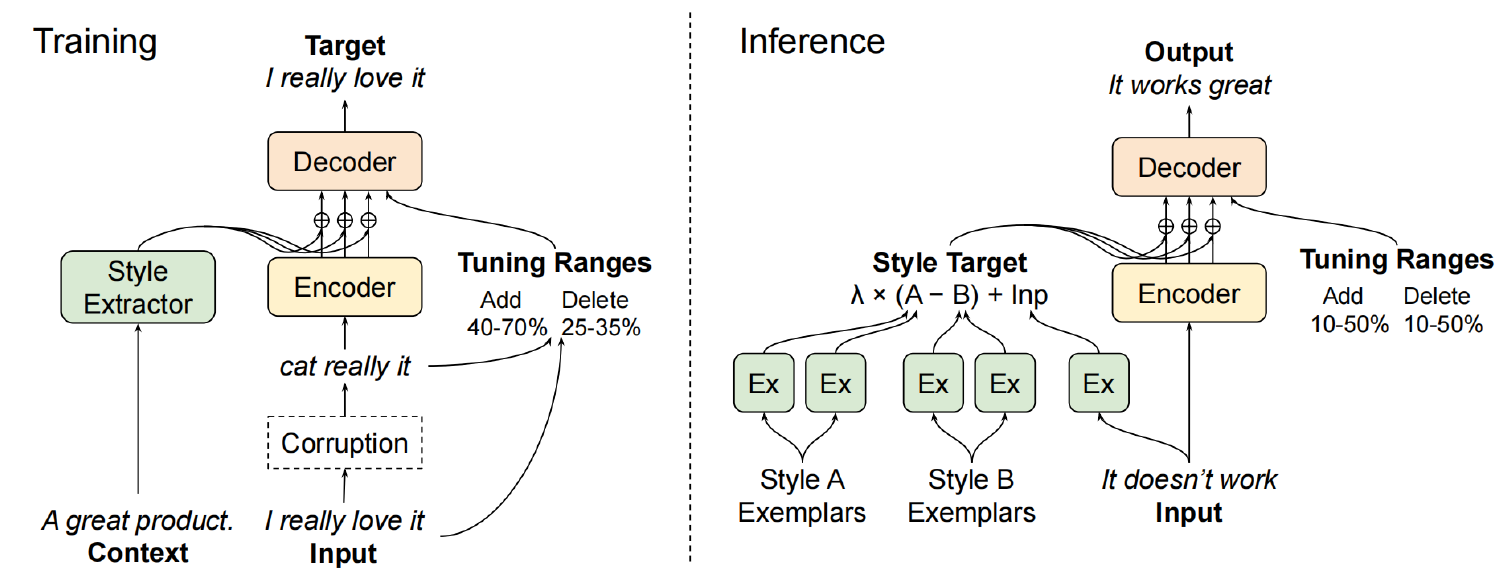}}
  \caption{
  A text emotion transfer model 
  from \cite{riley2021textsettr}.
  During training, the model restored a corrupted input by using a stable "style vector" derived from the preceding sentence to condition the reconstruction process. A new style vector allowed precise, targeted style modifications, or "targeted restyling," during inference by adjusting the direction and magnitude of the style shift relative to a baseline, using a few exemplar sentences to define the desired output style.
   }
   \label{fig: text editing}
\end{figure}

For example, in \cite{mohammadibaghmolaei2023tet}, Mohammadibaghmolaei et al. proposed a text emotion transfer technique based on masked language modeling and transfer learning, and a GPT-2 model underwent training to construct an initial sentence based on its altered sequences, allowing the model to perform efficiently even with limited emotion-annotated data. 
A novel text sentiment transfer methodology was proposed by Li et al. \cite{li2018delete} in which they employed a three-step process—Delete, Retrieve, and Generate. This approach, powered by unsupervised learning and neural sequence-to-sequence models, effectively altered sentiment while retaining content.
In \cite{jin2019imat}, Jin et al. introduced the IMaT model that constructed a pseudo-parallel corpus through semantic alignment, then applied a sequence-to-sequence model for attribute translation, refining this alignment across iterations. 
Wu et al. \cite{Wu2019} proposed a two-stage "Mask and Infill" methodology that significantly enhanced the performance of non-parallel text sentiment transfer. 
Following the "mask and infill" method, in \cite{malmi2020unsupervised}, Malmi et al. introduced an innovative unsupervised method using padded masked language models (MLMs) for sentiment transfer, using a padded MLM variant to avoid having to predetermine the number of inserted tokens. 
In \cite{yang2018unsupervised}, Yang et al. presented a technique leveraging language models as discriminators for unsupervised sentiment manipulation, enhancing stability and content fidelity in generated text. 
In \cite{shen2017style}, Shen et al. developed a technique for text sentiment transfer without parallel data, utilizing refined alignment of latent representations, which effectively separated content from style, allowing for sentiment modification by mapping sentences to a style-independent content vector, then decoding this vector into another style. 
Zhang et al. \cite{zhang2019emotional} employed GAN for sentiment transfer across different text domains, innovatively combining adversarial and reinforcement learning with a cross-domain sentiment transfer model, enhancing the ability to generate emotionally nuanced text while maintaining domain-specific content. 
In Fig. \ref{fig: text editing}, Riley et al. leveraged T5 to extract a style vector from preceding sentences and used it to provide extra conditioning for the decoder \cite{riley2021textsettr}. 
Huang et al. introduced a method \cite{huang2020cycle} based on Cycle-consistent Adversarial AutoEncoders, comprising LSTM autoencoders, adversarial style transfer networks, and cycle-consistent constraints, innovating unsupervised text style transfer. 
In \cite{luo2019towards}, Luo et al. proposed the Seq2SentiSeq model, incorporating sentiment intensity via a Gaussian kernel in the decoder, enhancing sentiment control. 
They trained the model using cycle reinforcement learning, which maintains the original message while changing emotional tone without requiring matched data pairs.

\subsection{Empathetic Dialogue Generation}
\label{sec:Empathetic Dialogue Generation}

Empathetic dialogue generation is a crucial aspect of creating more human-like and emotionally intelligent conversational agents. It goes beyond simply generating contextually relevant responses and focuses on incorporating emotional understanding and support into the generated dialogue.

A lot of researchers focused on how to generate responses with specific emotional tendencies or more empathy by LLMs.
For example, 
in \cite{li2024enhancing}, Li et al. employed the Emotional Chain-of-Thought (ECoT) technique, enhancing Large Language Models' capability for nuanced emotional text generation, focusing on human emotional intelligence alignment. 
Yang et al. \cite{yang2024enhancing} provided a Hybrid Empathetic Framework (HEF) that used SEMs as flexible enhancements to LLMs, implementing a two-stage emotion prediction strategy and an emotion cause perception strategy. 
In \cite{sun2023rational}, Sun et al. utilized the Chain of Emotion-aware Empathetic prompting (CoNECT) for better context understanding and emotional engagement. 
In \cite{lee2022does}, Lee et al. investigated GPT-3's capacity for empathetic dialogue generation, employing in-context learning in zero-shot and few-shot settings. 
Casas et al.  \cite{casas2021enhancing} introduced an empathic chatbot framework utilizing transformer-based language models for generating responses that recognized and adapted to the user's emotional state. 
In \cite{chen2023soulchat}, Chen et al. enhanced the empathetic responses of ChatGLM-6B by fine-tuning it with a specialized dataset comprising over 2 million multi-turn empathy conversation samples.

\begin{figure}[]
  \centerline{\includegraphics[width=1.0\linewidth]{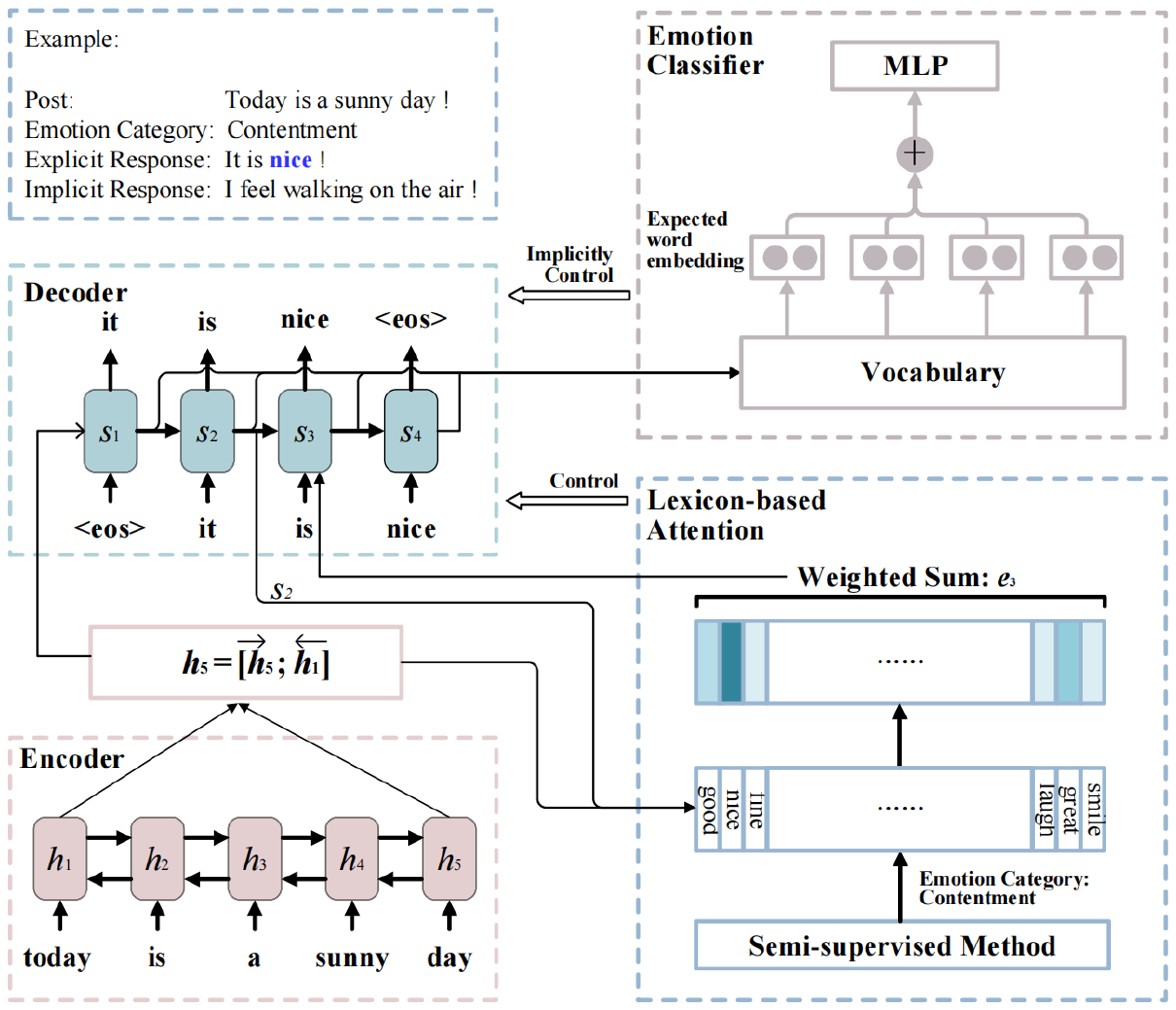}}
  \caption{A empathetic dialogue generation
  system 
  from \cite{song2019generating}. The system employed a bidirectional LSTM encoder to process the input text into a vector representation, which initialized a decoder. This decoder, guided by an emotion classifier and enriched with a lexicon-based attention mechanism, integrated emotional lexicon words seamlessly into the responses.
   }
   \label{figure10}
\end{figure}

Apart from LLMs, there are some studies on emotional or empathetic dialogue generation based on models like Seq2Seq model and conditional variational autoencoder (CVAE). 
For example, 
as shown in Fig. \ref{figure10}, Song et al. proposed an emotional dialogue system, EmoDS \cite{song2019generating}, that enhanced a Seq2Seq framework with a lexicon-based attention mechanism and an emotion classifier, generating dialogue responses that expressed specified emotions, either explicitly or implicitly. 
In \cite{zhou2018emotional}, Zhou et al. introduced an Emotional Chatting Machine (ECM) that integrated internal and external memory mechanisms along with emotion category embeddings. 
In \cite{kong2019adversarial}, Kong et al. proposed a model that effectively combined CGANs with either standard Seq2Seq or CVAE models to produce dialogue responses with the specified sentiment. 
Li et al. \cite{li2021dual} introduced a novel Dual-View CVAE model that synthesized emotional dialogue, changing the emotional expression of responses with higher content relevance. 
In \cite{xu2019generating}, Xu et al. proposed a framework that incorporated multi-task learning and dual attention mechanisms, effectively decoupling and processing content and emotional information from the input. 
Asghar et al. \cite{asghar2018affective} suggested an enhanced Seq2Seq model that incorporated three emotional strategies for the input, training, and inference processes, which was based on a designed dictionary with Valence, Arousal, and Dominance (VAD) scores. 
In \cite{colombo2019affect}, Colombo et al. employed a Seq2Seq architecture augmented by emotion embeddings and a VAD lexicon for word and sequence-level emotion modeling. It utilized an affect regularizer to favor emotionally charged words and an affect sampling method for generating emotionally relevant diverse responses. 
Huang et al. \cite{huang2018automatic} introduced a novel approach to integrate emotions into dialogue generation by appending an emotion token to the dialogue input or injecting the emotion directly into the decoder. 
In \cite{lin2022emotional}, Lin et al. proposed a model integrating Transformer and CVAE, with an emotion perception encoder and a BERT-based emotion classification model to embed emotional intelligence, enabling the generation of nuanced and contextually relevant empathetic responses.

In addition to the aforementioned methods, there have been several research efforts exploring the use of GANs to achieve empathetic dialogue generation.
For example, 
in \cite{wang2018sentigan, wang2019automatic}, Wang et al. developed SentiGAN, an innovative structure featuring numerous generators alongside a singular multi-class discriminator. 
This setup encouraged each generator to concentrate on crafting text samples that distinctly exhibited a designated sentiment label. 
Chen et al. \cite{chen2021emotional} utilized GAN with multiple classifiers to enhance emotional dialogue production, with an emotion discriminative model to align the generated dialogue's emotion with the intended one. 
Bi et al. \cite{bi2023diffusemp} utilized a diffusion model-based approach to generate empathetic responses, distinctive for its use of multi-grained control signals, incorporating communication mechanism, intent, and semantic frame as control levels, enabling nuanced guidance over the generated responses. 
In \cite{chen2020customizable}, Chen et al. proposed CTGAN for emotional text generation by incorporating emotion labels, allowing for the production of text that aligned with specific emotional tones within variable-length outputs.

\clearpage

\begin{table*}[t]
\centering
\caption{Literature on Generative Models for Text Emotion Transfer, Empathetic Dialogue Generation, in Textual Emotion Synthesis.}
\rotatebox{90}{  
    \begin{adjustbox}{max width=\textheight}
\begin{tabular}{|l|l|l|l|l|}
\hline
 Reference                         & Year     & Model   &Dataset &Performance         \\ \hline
                    Mohammadibaghmolaei et al. \cite{mohammadibaghmolaei2023tet}     & 2023  &  LLM    &ISEAR/TEC &Transfer strength figures/Content preservation: 0.8403-0.8967/Fluency: 164.9118-220.4884  \\ \hline 
                    Li et al. \cite{li2018delete}     &  2018  &  Seq2Seq    &YELP/CAPTIONS/AMAZON &Human evalution/BLEU: 11.8, 17.1, 27.1/Classifier Accuracy: 95.4, 96.8, 70.3       \\ \hline 
                    Jin et al. \cite{jin2019imat}     &  2019 &Seq2Seq   &YELP/FORMALITY &Human evalution/Accuracy: 95.90, 72.07/BLEU: 22.46, 38.16/PPL: 14.89, 32.63       \\ \hline
                    Wu et al. \cite{Wu2019}     &  2019 &LLM   &YELP/AMAZON &Human evalution/ACC: 97.3, 84.5/BLEU: 15.9, 32.1          \\ \hline
                    Malmi et al. \cite{malmi2020unsupervised}    & 2020   &LaserTagger  &YELP &BLEU: 14.5, 15.3/ACC: 40.9, 49.6           \\ \hline 
                    Yang et al. \cite{yang2018unsupervised} & 2018 &Encoder-Decoder LM &YELP &ACC: 90.0, 85.4/BLEU: 22.3, 13.4/$PPL_{X}$: 48.4, 32.8/$PPL_{Y}$: 61.6, 40.5 \\ \hline 
                    Shen et al. \cite{shen2017style}    & 2017  &AE &YELP &Accuracy:78.4/Human evaluation (sentiment: 62.6,fluency: 2.8,overall transfer quality: 41.5)             \\ \hline  
                    Zhang et al. \cite{zhang2019emotional}    & 2019   &Seq2Seq GAN    &YELP/AMAZON &Sentiment Transfer Strength: 81.7, 69.2/Cosine Similarity: 95.0, 92.2/Word Overlap: 83.9, 75.4          \\ \hline  
                    Luo et al. \cite{luo2019towards}      & 2019 &Seq2Seq  &YELP &BLEU: 32.5, 10.3/MAE: 0.13/MRRR: 0.78/PPL: 35.1/Avg human evaluation: 3.96        \\ \hline  
                    Reif et al. \cite{reif2022recipe}   & 2022  &LLM  &YELP/GYAFC &Human evalution figures/ACC: 90.6/BLEU: 10.4/PPL: 79         \\ \hline 
                    Yi et al. \cite{yi2021text}   & 2021  &Encoder-Decoder &YELP &ACC: 90.8/BLEU: 26.3/Cos: 96/PPL: 109/GM: 14.83          \\ \hline 
                    Li et al. \cite{li2020dgst}   & 2020  &Dual-Generator Network     &YELP/IMDb &ACC: 88.0, 70.1/BLEU: 54.5, 70.2      \\ \hline 
                    Huang et al. \cite{huang2020cycle}    & 2020  &AAE  &YELP &Tranfer: 86.9\%/BLEU: 22.51/PPL: 21.6/RPPL: 57.0        \\ \hline 
                    Riley et al. \cite{riley2021textsettr}     & 2021 &LLM &AMAZON &ACC: 73.7, 54.9/Content preservation: 34.7, 55.8/Sentiment: 2.5/Preservation: 2.6/Fluency: 4.0         \\ \hline  
                    Sancheti et al. \cite{sancheti2020reinforced}   & 2020   &Seq2Seq  &YELP &BLEU: 0.153,0.088/Accuracy: 0.922, 0.744/Human evaluation        \\ \hline 
                    Chawla et al. \cite{chawla2020semi}   & 2020  &Encoder-Decoder LM  &GYAFC/YELP/AMAZON &ACC: 86.2, 68.9/BLEU: 14.1, 28.6       \\ \hline

                    Li et al. \cite{li2024enhancing}     &  2024  &LLM  &IEMOCAP/DailyDialog/Empathetic-Dialogues/ESConv/PENS & EGS: 36.02, 36.42, 36.21, 36.70, 33.48         \\ \hline
                    Yang et al. \cite{yang2024enhancing}     &  2024 &LLM   &EmpatheticDialogue dataset &Acc: 45.63/Distinct-1: 42.23/Distinct-2: 80.08    \\ \hline
                    Sun et al.  \cite{sun2023rational}    &  2023 &LLM  &EmpatheticDialogue dataset &Human evaluation/PPL: 18.86/ACC: 53.44          \\ \hline
                    Lee et al. \cite{lee2022does}    & 2022 &LLM  &EmpatheticDialogue dataset &Human evaluation/EMOACC: 0.1683/Explorations: 0.4970/Interpretations: 0.2780           \\ \hline 
                    Casas et al. \cite{casas2021enhancing}    & 2021   &LLM &DD/ED dataset &Emotion Reflection: 0.465/Emotional: 0.487/Empathy Score: 0.443/PPL: 149.8          \\ \hline  
                    Chen et al. \cite{chen2023soulchat}    & 2023   &LLM  &SoulChat-Corpus/SMILECHAT &BLEU: 33.78, 20.07,12.86, 8.52/ROUGE: 31.47, 8.92, 26.57/Emp: 1.84/Human evaluation         \\ \hline  
                    Liu et al. \cite{liu2022empathetic}     & 2022  &LLM  &EmpatheticDialogues dataset &Emotion accuracy: 0.5262/Perplexity: 13.57/Dist-1: 2.04/Dist-2: 11.68          \\ \hline  
                    Qian et al. \cite{qian2023harnessing}    & 2023  &LLM  &EmpatheticDialogues dataset &Dist-1: 2.96/Dist-2: 18.29/BERTscore: 0.8803, 0.8816, 0.8774/BLEU-2: 9.37/BLEU-4: 3.26/Human evaluation        \\ \hline 
                    Song et al. \cite{song2019generating}   & 2019 &Seq2Seq  &STC/NLPCC &Embedding score: 0.634, 0.451, 0.435/BLUE: 1.73/Dist-1: 0.0113/Dist-2: 0.0867/emotion-a: 0.810/emotion-w: 0.687       \\ \hline 
                    Zhou et al. \cite{zhou2018emotional}    & 2018 &Seq2Seq &STC/NLPCC &Perplexity: 61.8/Accuracy: 0.773/Human evalution          \\ \hline 
                    Kong et al. \cite{kong2019adversarial}    & 2019 &CGAN Seq2Seq CAVE  &MojiTalk dataset &PPL: 69.54/ Sentiment Acc: 78.8, 78.9/Quality: 3.9       \\ \hline 
                    Li et al. \cite{li2021dual}  & 2021  &CAVE  &NLPCC2017/Weibo dataset/MojiTalk dataset &PPL: 25.70, 23.60, 33.7/Dist-1: 0.109, 0.017, 0.105/Dist-2: 0.400, 0.225, 0.517/Human evalution        \\ \hline 
                    Xu et al. \cite{xu2019generating}   & 2019 &CVAE Seq2Seq &NLPCC2017 &PPL: 34.6/Accuracy: 0.637/Dist-1: 0.3315/Dist-2: 0.7900/Dist-3: 0.9023/Human evalution          \\ \hline 
                    Asghar et al. \cite{asghar2018affective}  & 2018  & Seq2Seq   &Cornell Movie Dialogs Corpus   &Syntactic Coherence: 1.76/Naturalness: 1.09/Emotional Appropriateness: 1.10     \\ \hline 
                    Colombo et al. \cite{colombo2019affect}  & 2019  & Seq2Seq &OpenSubtitles2018/Cornell &Dist-1: 0.0406, 0.0305/Dist-2: 0.2030, 0.1431/BLEU: 0.0140, 0.110/Hyper-parameter optimization   \\ \hline 
                    Huang et al. \cite{huang2018automatic} & 2018  &Seq2Seq &CBET/OpenSubtitles dataset  &Average Acc: 76.69, 75.91, 78.46        \\ \hline
                    Lin et al. \cite{lin2022emotional}   & 2022&  CVAE &EmpatheticDialogues dataset &PPL: 19.6/Diversity: 0.0208, 0.1404, 0.3976/Human evaluation     \\ \hline
                    Firdaus et al. \cite{firdaus2020emosen}   & 2020&CVAE &SEMD &PPL: 34.8/Sentiment Acc:0.85/Emotion Acc: 0.80/Dist-1: 0.0203/Dist-2: 0.0520/Human evaluation        \\ \hline
                    Majumder et al. \cite{majumder2020mime}   & 2020 &Encoder-Decoder  &EmpatheticDialogues dataset &BLEU: 2.98/Human evaluation/Preference test      \\ \hline
                    Sabour et al. \cite{sabour2022cem}    & 2022 &Encoder-Decoder   &EmpatheticDialogues dataset &PPL: 35.60/Dist-1: 0.66/Dist-2: 2.99/Acc: 39.11/Human evaluation    \\ \hline
                    Li et al. \cite{li2019reinforcement}   & 2019 &Encoder-Decoder &NLPCC2017 &PPL: 62.2, 61, 61.4/Accuracy: 0.871, 0.870, 0.869/Human evaluation        \\ \hline
                    Wang et al. \cite{wang2019automatic}   & 2019  &GAN &MR/BR/CR/Emotional tweet conversation &Sentiment Acc: 0.885, 0.841, 0.803/Novelty: 0.395, 0.427, 0.549/Diversity: 0.741, 0.713, 0.708/Human evaluation      \\ \hline
                    Chen et al. \cite{chen2021emotional}   & 2021  &GAN &NLPW/XHJ &Acc: 0.701-0.870/Fluency score: 6.300-11.33/Human evaluation      \\ \hline
                    Bi et al. \cite{bi2023diffusemp}   & 2023 &Diffusion model &EmpatheticDialogues dataset &BERTScore: 0.5205/MIScore: 626.92/Acc: 92.36, 84.24/F1-SF: 52.79/Dist-1, 2, 4: 2.84, 29.25, 73.45/Self-BLEU: 1.09   \\ \hline
                    Li et al. \cite{li2020emotional}   & 2020 &GAN Seq2Seq  &NLPCC2017 &PPL: 62.8/Acc: 0.735/Distinct: 0.0287, 0.3062/Human evaluation      \\ \hline
                    Chen et al. \cite{chen2020customizable}    & 2020 &GAN &Yelp restaurant reviews/Amazon reviews/Film review/Obama speech &Similarity: 0.1-0.2/Sentiment figures/Average Acc: 0.56      \\ \hline
                    Gu et al. \cite{gu2024multilevel}    & 2024 &LLM &EmpatheticDialogues dataset &Dist-1: 2.98/Dist-2: 18.38/B-2: 7.64/B-4: 2.57/PBERT: 87.29/RBERT: 88.04/FBERT:87.63/Human evaluation(Empaythy): 4.17      \\ \hline
                    Chen et al. \cite{ chen2024cause}    & 2024 &LLM &EmpatheticDialogues dataset &Acc: 52.73/Dist-1: 2.96/Dist-2: 19.52/BLEU-2: 10.54/BLEU-4: 5.17      \\ \hline
\end{tabular}%
\label{text table}
\end{adjustbox}
}
\end{table*}

\clearpage

\section{Evaluation Metric}
\label{sec:Evaluation Metric}
In the field of human emotion synthesis, several commonly used evaluation metrics are employed to assess the quality of generated content across the three modalities of facial images, speech, and text. 
These metrics provide a comprehensive assessment of the synthesized emotions from various perspectives. While some evaluation indicators are universal and applicable to all modalities, others are exclusive to specific modalities, as illustrated in Table~\ref{table: metrics}.

For common evaluation indicators, there are usually two forms: user study and accuracy. 
User studies are a qualitative evaluation method that assesses the quality of generated content from aspects such as clarity, naturalness, and emotional authenticity. These studies involve gathering feedback and opinions from human participants to gauge their perception and experience of the synthesized emotions.
For example, in tasks like face reenactment, where emotional authenticity is critical, researchers conduct a "Real vs. Fake" perceptual study on platforms like Amazon Mechanical Turk (AMT) to evaluate the outputs." 
In the field of TTS and voice conversion, researchers \cite{tang2023emomix, um2020emotional, oh2024durflex} use subjective evaluation metrics like Mean Opinion Score (MOS) and Similarity Mean Opinion Score (SMOS) to assess naturalness and emotional similarity. 
Accuracy is another important evaluation metric for human emotion synthesis.
For example, in face reenactment, researchers use classifiers to assess generated facial images. 
Higher accuracy in expression recognition reflects higher accuracy in expression translation by the models \cite{shao2021wp2}. 
Similarly, \cite{zhang2024emotalker} utilizes an emotion classifier network from EVP \cite{ji2021audio} to measure the emotion accuracy of face manipulation.

In addition to the aforementioned universal evaluation metrics, different modalities in the field of emotion synthesis have their own specific and widely used evaluation indicators. 
These indicators target the unique attributes and synthesis goals of each modality, providing a more fine-grained and specialized assessment perspective.
For example, in facial 
emotion synthesis, PSNR (Peak Signal-to-Noise Ratio) \cite{song2018geometry,guo2024gaussianpu} quantifies image quality by comparing compressed images to their originals. It is calculated using the logarithm of the ratio between maximum pixel value and mean squared error, with higher values indicating better quality. 
SSIM (Structural Similarity Index) \cite{shao2021wp2} improves upon PSNR by evaluating image similarity based on luminance, contrast, and structure, focusing on perceived quality and structural integrity. 
FID (Fréchet Inception Distance) \cite{zeng2020realistic} measures the similarity between sets of images by comparing feature vectors, with lower scores indicating higher quality and diversity in generated images.
In speech emotion synthesis, MCD (Mel Cepstral Distortion) \cite{tang2023emomix,cornille2022interactive} objectively measures spectral similarity between reference and generated mel-spectrograms, providing quantitative feedback on emotional voice synthesis accuracy. 
F0 RMSE (F0 Root Mean Square Error) \cite{wu2019end,im2022emoq} evaluates the accuracy of the fundamental frequency contour in synthesized speech compared to the reference. 
Lower values indicate higher pitch accuracy, contributing to perceived naturalness and expressiveness. 
In textual emotion synthesis, the BLEU (Bilingual Evaluation Understudy) score \cite{li2018delete,Wu2019}, borrowed from machine translation evaluation, can measure lexical similarity between generated and reference texts, indicating how well the generated text maintains desired linguistic properties while altering emotional tone. 
PPL (Perplexity) \cite{jin2019imat} measures a language model's prediction accuracy, reflecting its ability to produce coherent and fluent text. Lower PPL suggests the generated text more closely mirrors human language patterns.

\renewcommand{\arraystretch}{1.3}
\begin{table}[ht]
\centering
\caption{Evaluation Metrics in Human Emotion Synthesis.}
\begin{tabular}{|>{\centering\arraybackslash}m{1.5cm}|>{\centering\arraybackslash}m{1.2cm}|>{\arraybackslash}m{5cm}|}
\hline
Modalities & Metrics & Description \\ \hline
\multirow{2}{*}[-1ex]{Common} & User study & Assesses the quality and realism of synthesized emotion contents based on scoring by human participants. \\ \cline{2-3} 
 & Accuracy & Measures the percentage of correct predictions in emotion classification tasks. \\ \hline
\multirow{3}{*}[-7ex]{\begin{tabular}[c]{@{}c@{}}Facial\\Emotion\\Synthesis\end{tabular}} & PSNR & Quantifies the similarity between the synthesized and original images, where higher values indicate better quality. \\ \cline{2-3} 
 & SSIM & Assesses the perceived visual quality by comparing structural information between the original and generated images. \\ \cline{2-3} 
 & FID & Evaluates the distribution difference between real and synthesized images by comparing deep feature representations. \\ \hline
\multirow{2}{*}[-0ex]{\begin{tabular}[c]{@{}c@{}}Speech\\Emotion\\Synthesis\end{tabular}} & MCD & Measures the distortion between the synthesized and reference speech in the cepstral domain. \\ \cline{2-3} 
 & F0 RMSE & Quantifies the difference in pitch between synthesized and real speech. \\ \hline
\multirow{2}{*}[-1ex]{\begin{tabular}[c]{@{}c@{}}Textual\\Emotion\\Synthesis\end{tabular}} & BLEU & Computes the lexical similarity between synthesized and reference text. \\ \cline{2-3} 
 & PPL & Measures the fluency of generated text, with lower values indicating more predictable and coherent sentences. \\ \hline
\end{tabular}
\label{table: metrics}
\end{table}

\section{Discussion}
\label{sec:Discussion}
\subsection{Major Findings}
Existing methods in generative technology for human emotion synthesis have made substantial progress across multiple modalities, including facial images, speech, and text. Each modality benefits from distinct approaches and these developments not only enhance the perceived emotional intelligence of systems but also push the boundaries of how machines can generate and interpret nuanced emotional states.

In facial emotion synthesis, GAN-based methods used to be the mainstream methods. However, in recent years, DM-based methods have also gained considerable attention and made significant progress. Specifically, some classical GAN-based models, such as StarGAN and StyleGAN, excel at altering the emotional expressions of faces but face challenges when attempting to generate more subtle or complex emotional states. Moreover, generating mixed emotions—that is, facial expressions that convey a strong contrast of emotions like happiness and sadness—still lacks finer control.
Comparatively, 
DM generates images through progressive denoising, which avoids the problem of model collapse that GANs may have when generating facial expressions. 
It also shows strong adaptability in generating mixed emotions, better capturing subtle expression changes and emotional nuances, and can handle multiple combinations of emotions in facial expressions.

Speech emotion synthesis achieved similar progress through the adaptation of GANs and Seq2Seqs like Tacotron \cite{wang2017tacotron}, where emotional intonation is introduced into synthesized speech. 
These models produce more natural and expressive speech by adjusting vocal elements like pitch, rhythm, and tone.
However, recent research has also incorporated AEs, and DMs to further improve the emotional depth and expressiveness of synthesized speech. 
Specifically, AEs are used to disentangle emotional features from speaker identity, enabling more flexible emotion transfer while preserving the naturalness of speech. DMs, with their capacity for modeling complex data distributions, offer promising results in generating emotional speech with high fidelity and more controlled variations in prosody.   

Textual emotion synthesis 
has increasingly leveraged LLMs and Seq2Seq architectures, utilizing mechanisms like sentiment control and emotional valence modulation to produce emotionally resonant content \cite{sorin2023large}. 
These systems are often employed in applications such as empathetic chatbots and emotionally responsive dialogue systems. Despite their effectiveness, generating responses with emotional depth that appropriately reflect varying levels of empathy, sympathy, or other complex emotional tones based on user inputs remains a challenge. 
Current models still struggle with maintaining a balance between emotional expressiveness and conversational coherence, especially in response to ambiguous or contextually nuanced inputs. Moreover, understanding the contextual triggers for emotional responses and integrating them effectively into generative models will be an ongoing research area. 

In terms of evaluation metrics, 
quality assessment remains a complex, multidimensional task that integrates both subjective and objective indicators.
Subjective metrics typically involve human evaluation, which focus on capturing the emotional authenticity, resonance, and overall impression of the generated content. 
However, subjective evaluations can be time-consuming and costly, and they are influenced by individual biases and cultural differences.
On the other hand, 
objective metrics aim to quantify various aspects of emotion synthesis using automated computational methods, such as accuracy, PSNR, etc. 
They provide a scalable and reproducible way of assessment but may not fully capture the emotional nuances and human-perceived quality of the generated content.
Future research directions include developing more fine-grained subjective evaluation methods to better capture the subtle differences in emotional content and the complexity of human responses. At the same time, improving objective metrics to more accurately quantify various aspects of emotional expression and correlating them with human judgments is also an important area of research.

\subsection{Future Perspectives}
Advances in generative AI have revolutionized how we simulate human emotions, creating more authentic and nuanced emotional expressions. 
Looking ahead, several promising avenues for further research can be explored:

Firstly, 
combining the capabilities of different generative models such as GANs, Seq2Seqs, AEs, DMs, and LLMs holds promise for further enhancing the quality of generated outputs. Each model has its unique strengths and limitations, and intelligently integrating them can compensate for the shortcomings of individual models, enabling more accurate and realistic emotion synthesis.
By designing innovative hybrid architectures that leverage the strengths of each model, more powerful and comprehensive emotion synthesis systems can be developed. These hybrid models can seamlessly transition between different modalities, generating emotionally consistent and complementary outputs. 
For instance, a model combining DMs with the insights of Seq2Seq can generate high-quality emotional speech with appropriate facial expressions and lip synchronization.

Secondly, the horizon of emotion synthesis is not limited to common modalities like facial images, speech, and text. 
As generative models continue to evolve, we may see multimodal human emotion synthesis results \cite{baltruvsaitis2018multimodal,hou2024visualrwkv} in the future, including modalities beyond those mentioned in this paper, such as gesture \cite{nyatsanga2023comprehensive} and physiological signals like EEG and ECG. 
In addition, emerging cross-modal generative models, such as text-to-image (T2I), text-to-video (T2V), and even text-to-3D, are poised to expand the creative and interactive potential of emotion synthesis. 
T2I models can generate imagery that resonates emotionally with written narratives, producing visuals that reflect subtle emotional undertones, while T2V models can bring stories to life by translating emotional content into animated, visually expressive scenes that engage audiences on a deeper emotional level. 
Moreover, as the technology matures, the potential for converting image-based emotions into sound (e.g., generating soundscapes that mirror the mood of a visual scene) opens up new dimensions for immersive experiences in fields like VR \cite{papoutsi2021virtual} and interactive entertainment.

Thirdly, the integration of generative models with edge devices—from server-based processing to smart terminals—marks a pivotal shift in the accessibility and application of emotion synthesis \cite{cao2020overview}. 
As the computational power of edge devices continues to grow, there is a growing potential for real-time emotion generation and synthesis directly on devices such as smartphones, wearables, and VR headsets. 
This transition from centralized server processing to edge computing opens up a wide range of applications, enabling personalized, on-the-fly emotional interactions. 
For instance, smart devices can analyze users' facial expressions, voice tone, or even physiological signals, generating responsive emotional content that adapts to the user's immediate emotional state, location, or context. 
Additionally, as AI models become more efficient, smaller-scale devices like wearables or even IoT sensors \cite{dian2020wearables} can incorporate emotion-aware interactions, enhancing user experience in a range of industries—from healthcare, where emotion synthesis can assist in mental health monitoring and intervention, to retail, where it can personalize consumer experiences in-store or online. 
In this new paradigm, generative models are poised to provide immediate, adaptive emotional content, offering users a deeper connection to the digital world.

Fourthly, 
human emotion synthesis holds the potential to drive profound transformations across a variety of industries and can revolutionize domains such as digital entertainment and filmmaking \cite{izani2024impact, channa2024original}. 
In the realm of digital entertainment, emotion synthesis lays the foundation for highly immersive experiences, enabling virtual characters and environments to express emotions with a level of authenticity rivaling that of human actors. By precisely generating emotional nuances in facial expressions, vocal tones, and body language, these technologies can elevate interactive media to new heights.
In film production, AI-driven tools are already being employed to infuse emotional depth into character performances, achieving more dynamic and expressive storytelling that transcends the limitations of traditional physical acting. 
Moreover, the combination of AIGC video and emotion synthesis opens up new avenues for creation, empowering filmmakers to craft content with emotionally evocative visual and auditory cues. 
This presents opportunities for real-time emotional adjustments within films, where characters' emotional arcs can be dynamically altered based on audience feedback or narrative shifts, further immersing viewers in the experience \cite{sun2023application}.
Through these advancements, the industry is entering a new chapter where the integration of emotion synthesis will unleash vast creative potential.

\section{Conclusion}
\label{sec:Conclusion}
This review presents a detailed investigation of current generative technology for human emotion synthesis across various modalities, including facial images, speech, and text. It reveals how different genrative models, ranging from well-established approaches like AEs and GANs to emerging techniques such as DMs and LLMs, are capable of generating complex emotional expressions with remarkable depth and subtle nuances.

In Section \ref{sec:introduction}, we introduce the background of human emotion synthesis, a pivotal area of research within affective computing. With the growing sophistication of generative models, which possess advanced data modeling and multi-modal generation capabilities, new avenues for emotion synthesis are emerging.  
However, there is currently a lack of comprehensive reviews on this subject. To address this gap, we present the first survey of generative technology for human emotion synthesis, building upon existing literature and filling the void in current research.
In Section \ref{sec:Review Methodology}, we outline our rigorous literature screening strategy, which allowed us to systematically collect and classify key research in the field of generative models for emotion synthesis. In Sections \ref{sec:Emotion Model} to \ref{sec:Databases}, we highlight the gaps between previous reviews and our survey, demonstrating the novelty and significance of our contribution. We also provide an overview of emotion models and mathematics of generative models, as well as commonly used datasets, offering a deeper understanding of the latest advancements in this interdisciplinary field.

Sections \ref{sec:Facial Emotion Synthesis} to \ref{sec:Textual Emotion Synthesis} provide
a detailed discussion of the latest research on human emotion synthesis based on facial images, speech, and text. We categorize the specific tasks under each modality, discuss the applications of different generative models, and summarize the performance of existing works in comprehensive tables.
We classify specific tasks within each modality, analyze the distribution of different models across these tasks, and discuss their strengths and limitations in various contexts. For each modality, we summarize the application models and performance of existing works in comprehensive tables.
Finally, in Sections \ref{sec:Evaluation Metric} and \ref{sec:Discussion}, 
we summarize the commonly used evaluation metrics for emotion synthesis and discuss the current state and future development trends in this field, providing insights into the challenges and opportunities faced by emotion synthesis.

In summary, this review highlights the transformative potential of existing generative models in shaping the future of human emotion synthesis. These advancements will not only enhance the granularity and authenticity of synthesized emotions but also usher in a new era where machines can resonate with the subtleties of human emotions, fostering deeper and more empathetic connections.


%





\ifCLASSOPTIONcaptionsoff
  \newpage
\fi

\end{document}